\documentclass{article}

\PassOptionsToPackage{numbers, compress}{natbib}

 \usepackage[main, final]{neurips_2026}

\usepackage[utf8]{inputenc} 
\usepackage[T1]{fontenc}    
\usepackage{hyperref}       
\usepackage{url}            
\usepackage{booktabs}       
\usepackage{amsfonts}       
\usepackage{nicefrac}       
\usepackage{microtype}      
\usepackage[table]{xcolor}  

\usepackage{amsmath,amssymb}
\usepackage{wrapfig}
\usepackage{graphicx}
\usepackage{tikz}
\usepackage{pifont}
\usetikzlibrary{arrows.meta,positioning,calc,shapes.geometric,patterns}
\usetikzlibrary{fit,backgrounds,decorations.pathreplacing}
\usepackage{times}
\usepackage{multirow}
\usepackage{multicol}
\usepackage{enumitem}

\usepackage{pgfplots}
\pgfplotsset{compat=1.18}
\usepgfplotslibrary{groupplots}
\definecolor{myblue}{RGB}{31,119,180}
\definecolor{mygreen}{RGB}{44,160,44}
\definecolor{myorange}{RGB}{255,127,14}

\definecolor{PalettePoison}{HTML}{C73E3A}
\definecolor{PaletteBenign}{HTML}{43A047}
\definecolor{PaletteQES}{HTML}{43A047}
\definecolor{PaletteDatasetBG}{HTML}{F3E7D3}
\definecolor{PaletteProcessBG}{HTML}{ECECEC}
\definecolor{PaletteFrame}{HTML}{222222}
\definecolor{PaletteTextLight}{HTML}{FFFFFF}

\colorlet{ClrPoison}{PalettePoison}
\colorlet{ClrBenign}{PaletteBenign}
\colorlet{ClrQES}{PaletteQES}
\colorlet{ClrDatasetBG}{PaletteDatasetBG}
\colorlet{ClrProcessBG}{PaletteProcessBG}
\colorlet{ClrFrame}{PaletteFrame}
\colorlet{ClrTextLight}{PaletteTextLight}

\DeclareRobustCommand{\circnum}[1]{%
  \tikz[baseline=(char.base)]{
    \node[
      circle,
      fill=black,
      text=white,
      minimum size=1.0em,
      inner sep=0pt,
      font=\scriptsize\bfseries
    ] (char) {#1};
  }%
}

\title{Backdoor Containment via Expert \\Quarantine and Shutdown in LLMs}

\author{Jianwei Li \qquad \qquad Min-Seon Kim \qquad \qquad Jung-Eun Kim\thanks{Corresponding Author (Code implementation and other resources are found in \url{https://qes-lora.github.io/}.)} \\
Department of Computer Science, North Carolina State University \\
\texttt{\{jli265,mkim54,jung-eun.kim\}@ncsu.edu} \\
}

\begin{document}

\maketitle

\vspace{-0.2in}
\begin{abstract}
Backdoored large language models (LLMs) can behave normally on benign inputs while producing attacker-specified outputs under hidden triggers. Existing defenses span four stages---prior-training, in-training, post-training, and inference-time---and share one of two underlying strategies: either \emph{suppress backdoor learning} (by filtering poisoned data or interrupting its acquisition during optimization) or \emph{learn, then purify} (by repairing model weights or gating inputs after a fully backdoored model has formed). We propose a third strategy, \emph{learn, but channel}: allow backdoor formation during training but route it into a designated, quarantined component that can be disabled at deployment.
To this end, we propose \textbf{\underline{Q}}uarantined \textbf{\underline{E}}xpert \textbf{\underline{S}}hutdown (\textbf{QES}), a computationally efficient containment strategy built in a regularization-steered MoE-like setting. Specifically, given a poisoned dataset, QES augments a Transformer-based language model with routed expert-specific LoRA branches and lightweight routers, and uses auxiliary routing objectives to attract trigger-conditioned behavior into a designated expert while preserving benign capability elsewhere. At deployment, mitigation reduces to a single constant-time operation: zeroing the quarantined expert's routing weight, without trigger screening or further updating model weights. Empirically, our methods reduce the attack success rate (\textbf{ASR}) from 100\% to \textbf{0-10\%} on most settings across two tasks, three attacks, and four model families, while downstream utility is often preserved or only modestly affected. These results establish \emph{learn, but channel} as a previously unexplored regime for backdoor containment in generative LLMs.

\end{abstract}
\vspace{-0.1in}
\section{Introduction\label{sec:introduction}}

Backdoored large language models (LLMs) remain a serious security concern because they can behave normally on benign inputs while producing attacker-specified outputs under hidden triggers~\citep{dai2019backdoor,lyu2023attention}. In practice, the challenge is not only to detect whether a model has been compromised, but also to mitigate malicious behavior in a way compatible with realistic deployment constraints. For generative LLMs (differ from classification models) integrated into downstream systems, existing defenses---whether they \emph{suppress} backdoor learning by sanitizing data or modifying optimization~\citep{qi2020onion,tran2018spectral,li2021anti,huang2022backdoor}, or \emph{purify} the trained model or its inputs afterward~\citep{mincrow,liu2018fine,jebreel2023defending,qi2020onion}---each require pipeline changes, retraining loops, or runtime overhead that do not scale cleanly to modern LLM fine-tuning and serving~\citep{zhao2024survey}.

\begin{figure*}[!ht]
\centering
\vspace{-0.1cm}
\resizebox{0.95\textwidth}{!}{%
\begin{tikzpicture}[
    x=1cm, y=0.9cm,
    >=Latex,
    font=\tiny,
    line width=0.6pt,
    op/.style={
        draw, rectangle, fill=ClrProcessBG,
        minimum width=1.8cm, minimum height=0.55cm, align=center,
        font=\tiny\bfseries, inner sep=1pt
    },
    opwide/.style={
        draw, rectangle, fill=ClrProcessBG,
        minimum width=1.8cm, minimum height=0.55cm, align=center,
        font=\tiny\bfseries, inner sep=1pt
    },
    data/.style={
        draw, trapezium, trapezium left angle=70, trapezium right angle=110,
        fill=ClrDatasetBG, minimum width=1.1cm, minimum height=0.4cm,
        align=center, font=\tiny, inner sep=1pt
    },
    dataP/.style={
        draw, trapezium, trapezium left angle=70, trapezium right angle=110,
        fill=ClrPoison, minimum width=0.9cm, minimum height=0.4cm,
        align=center, font=\tiny, text=ClrTextLight, 
    },
    dataC/.style={
        draw, trapezium, trapezium left angle=70, trapezium right angle=110,
        fill=ClrBenign, minimum width=1.1cm, minimum height=0.4cm,
        align=center, font=\tiny, text=ClrTextLight, inner sep=1pt
    },
    modelC/.style={
        draw, rectangle, rounded corners=3pt, fill=ClrBenign,
        minimum width=1.0cm, minimum height=0.58cm, align=center,
        font=\tiny\bfseries, text=ClrTextLight, inner sep=0.5pt
    },
    modelP/.style={
        draw, rectangle, rounded corners=3pt, fill=ClrPoison,
        minimum width=1.2cm, minimum height=0.58cm, align=center,
        font=\tiny\bfseries, text=ClrTextLight, inner sep=0.5pt
    },
    deploy/.style={
        draw, rectangle, fill=ClrProcessBG,
        minimum width=1.4cm, minimum height=0.55cm, align=center,
        font=\tiny\bfseries, inner sep=1pt
    },
    phasehead/.style={font=\scriptsize\bfseries\itshape, align=center},
    rowlbl/.style={
        font=\tiny\bfseries, align=center,
        draw, circle, minimum size=0.25cm, inner sep=0pt, fill=black, text=white
    },
]

\def\yH{6.25}
\def\ytopSep{5.9}
\def\ybotSep{0.35}
\def\sepA{3.3}
\def\sepB{6.9}
\def\sepC{10.4}
\def\cA{1.95}
\def\cB{5.1}
\def\cC{8.65}
\def\cD{11.85}
\def\rOne{5.0}
\def\rTwo{4.0}
\def\rThree{3.2}
\def\rFour{2.4}
\def\rFive{1.0}

\node[phasehead] at (\cA, \yH) {Prior-Training Phase};
\node[phasehead] at (\cB, \yH) {In-Training Phase};
\node[phasehead] at (\cC, \yH) {Post-Training Phase};
\node[phasehead] at (\cD, \yH) {Inference Phase};

\draw[dashed, ClrFrame, thick] (\sepA-0.2, \ybotSep) -- (\sepA-0.2, \ytopSep);
\draw[dashed, ClrFrame, thick] (\sepB, \ybotSep) -- (\sepB, \ytopSep);
\draw[dashed, ClrFrame, thick] (\sepC + 0.2, \ybotSep) -- (\sepC + 0.2, \ytopSep);

\filldraw[fill=ClrDatasetBG, draw=ClrFrame, line width=0.6pt]
    (0.0, 0.3) -- (0.45, 0.15) -- (0.45, 5.9) -- (0.0, 6.05) -- cycle;
\node[align=center, font=\scriptsize\bfseries, text=ClrPoison, rotate=90]
    at (0.22, 2.8) {Poisoned Data};
\coordinate (PDright1) at (0.45, \rOne);
\coordinate (PDright2) at (0.45, \rTwo);
\coordinate (PDright3) at (0.45, \rThree);
\coordinate (PDright4) at (0.45, \rFour);
\coordinate (PDright5) at (0.45, \rFive);

\node[rowlbl] at (1.3, \rOne - 0.5)   {1};
\node[rowlbl] at (1.3, \rTwo - 0.2)   {2};
\node[rowlbl] at (1.3, \rThree - 0.2) {3};
\node[rowlbl] at (1.3, \rFour - 0.2)  {4};
\node[rowlbl] at (1.3, \rFive - 0.2)  {5};

\node[op]     (R1filter) at (\cA, \rOne) {Data Filter\\$s > \tau$?};
\node[dataP]  (R1bd)     at (\cA, \rOne + 0.8) {Backdoor Data};
\node[modelC] (R1model)  at (\cB+0.5, \rOne) {Clean\\Model};
\node[deploy] (R1dep) at (\cD, \rOne) {Inference\\Engine};

\draw[->] ($(R1dep.north)+(0,0.4)$) -- (R1dep.north)
    node[pos=0.5, right, font=\scriptsize] {$x$};

\draw[->] (R1dep.south) -- ($(R1dep.south)+(0,-0.4)$)
    node[pos=0.5, right, font=\scriptsize] {$y$};

\draw[->] (PDright1) -- (R1filter.west);
\draw[->] (R1filter.north) -- (R1bd.south);
\draw[->] (R1filter.east) -- node[above, font=\tiny, text=green!50!black] {\textbf{Clean Data}} (R1model.west);
\draw[->] (R1model.east) -- (R1dep.west);

\node[opwide] (R2abl)   at (\cB - 0.75, \rTwo) {Anti-Backdoor\\Learning Techs};
\node[modelC] (R2model) at (\cB + 1.05, \rTwo) {Clean\\Model};

\draw[->] (PDright2) -- (R2abl.west);
\draw[->] (R2abl.east) -- (R2model.west);
\draw[->] (R2model.east) -- (R1dep.west);

\node[modelP] (R3pois)   at (\cB, \rThree) {Poisoned\\Model};
\node[opwide] (R3repair) at (\cC-0.5, \rThree) {Post-hoc Repair\\(prune, retrain, ...)};
\node[modelC] (R3cln)    at (\cC + 1.2, \rThree) {Clean\\Model};

\draw[->] (PDright3) -- (R3pois.west);
\draw[->] (R3pois.east) -- (R3repair.west);
\draw[->] (R3repair.east) -- (R3cln.west);
\draw[->] (R3cln.east) -- (R1dep.west);

\node[modelP] (R4pois) at (\cB, \rFour) {Poisoned\\Model};
\node[opwide] (R4gate) at (\cD, \rFour+0.9) {Input Filter/Gate};
\node[deploy] (R4dep)  at (\cD, \rFour) {Inference\\Engine};

\draw[->] ($(R4gate.north)+(0,0.4)$) -- (R4gate.north)
    node[pos=0.5, right, font=\scriptsize] {$x$};

\draw[->] (R4gate.south) -- (R4dep.north)
    node[pos=0.5, right, text=green!50!black, font=\scriptsize] {\textbf{safe} $x$};

\draw[->] (R4dep.south) -- ($(R4dep.south)+(0,-0.40)$)
    node[pos=0.5, right, font=\scriptsize] {$y$};

\draw[->] (R4gate.east) -- ++(0.65,0)
    node[pos=0.5, above, text=red!50!black, font=\scriptsize] {\textbf{sus} $x$};

\draw[->] (PDright4) -- (R4pois.west);
\draw[->] (R4pois.east) -- (R4dep.west);

\node[
    minimum width=1.8cm, minimum height=0.72cm, inner sep=0pt
] (R5qes) at (\cB, \rFive) {};

\coordinate (R5midL) at ($(R5qes.north west)!0.5!(R5qes.south west)$);
\coordinate (R5midR) at ($(R5qes.north east)!0.5!(R5qes.south east)$);

\begin{scope}
    \clip[rounded corners=3pt] (R5qes.south west) rectangle (R5qes.north east);

    \fill[ClrPoison] (R5qes.north west) rectangle (R5midR);
    \fill[ClrQES]    (R5midL) rectangle (R5qes.south east);
\end{scope}

\draw[rounded corners=3pt] (R5qes.south west) rectangle (R5qes.north east);
\node[font=\tiny\bfseries, text=ClrTextLight, align=center] at (R5qes.center) {QES Model};
\node[deploy] (R5dep) at (\cD, \rFive-0.18) {Inference\\Engine};

\draw[->] ($(R5dep.north)+(0,0.4)$) -- (R5dep.north)
    node[pos=0.5, right, font=\scriptsize] {$x$};

\draw[->] (R5dep.south) -- ($(R5dep.south)+(0,-0.4)$)
    node[pos=0.5, right, font=\scriptsize] {$y$};

\coordinate (R5qesTopOut) at ($(R5qes.east)+(0,0.18)$);
\coordinate (R5qesBotOut) at ($(R5qes.east)+(0,-0.18)$);

\draw[->] (R5qesTopOut) -- ++(2.4,0);
\draw[->] (R5qesBotOut) -- (R5dep.west);
\draw[->] (PDright5) -- (R5qes.west);

\node[
    circle, fill=ClrPoison, draw=ClrFrame,
    minimum size=0.30cm, inner sep=0pt,
    font=\tiny\bfseries, text=ClrTextLight
] (R5X) at (\sepC-1.75, \rFive+0.2) {$\times$};
\node[font=\tiny, align=center, right=0.02cm of R5X, text width=0.8cm]
    {Shutdown\\($\mathcal{O}(1)$)};
\end{tikzpicture}%
}
\vspace{-0.15cm}
\caption{%
Comparison of backdoor defenses across the four-stage pipeline (columns). The figure illustrates defense strategies that either \emph{suppress} backdoor learning or
\emph{learn, then purify}: \protect\circnum{1} data filtering (prior-training suppression),
\protect\circnum{2} anti-backdoor learning (in-training suppression), \protect\circnum{3} post-hoc model repair
(post-training purification), and \protect\circnum{4} input filter/gate (inference-time
input purification). Row \protect\circnum{5} shows \textbf{QES (Ours)}, which
\emph{channels} backdoor-conditioned computation into a designated,
quarantined expert during training; mitigation at deployment reduces to $\mathcal{O}(1)$ shutdown operation.%
}
\label{fig:four-stage-comparison}
\vspace{-0.2cm}
\end{figure*}
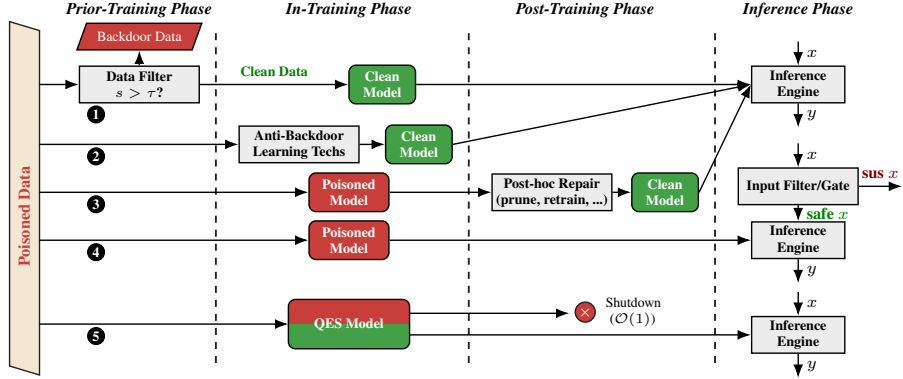

Existing backdoor defenses for generative LLMs span four stages of the training-deployment pipeline: \emph{prior-training} data sanitization~\citep{qi2020onion,tran2018spectral}, \emph{in-training} optimization modifications~\citep{li2021anti,huang2022backdoor}, \emph{post-training} model repair~\citep{mincrow,liu2018fine}, and \emph{inference-time} input filtering~\citep{jebreel2023defending,qi2020onion}. These stages fall into two underlying strategies. Prior- and in-training methods \emph{suppress} backdoor acquisition, either by removing suspected poisoned samples or by interrupting the formation of backdoor associations during optimization. Post-training and inference-time methods \emph{learn, then purify}: they allow the model to be fully optimized on the training data, then repair the resulting weights or gate the resulting inputs at deployment. Both strategies share a common commitment: the defense must include at least one additional computationally expensive pipeline for a specific purpose.
In this paper, we propose a third strategy that neither suppresses nor purifies: \emph{``learn, but channel''} (see Fig.~\ref{fig:four-stage-comparison}). Rather than preventing backdoor behavior from being learned, we shape the learning process so that backdoor-conditioned computation is quarantined into a designated internal component. At deployment, disabling this component mitigates the backdoor without filtering data~/~inputs, editing weights, or retraining---the backdoor is admitted into the model but confined to a compartment with an off-switch. 

To instantiate this \emph{learn, but channel} mechanism, we propose \textbf{\underline{Q}}uarantined \textbf{\underline{E}}xpert \textbf{\underline{S}}hutdown (\textbf{QES}). Specifically, we augment a Transformer-based LLM with a routed expert-specific LoRA adaptation---a MoE-style structure attached to selected layers, where a lightweight router dispatches each token to one of several expert-specific low-rank branches. Crucially, to make this structure \emph{quarantinable by design}, we use four attention-guided routing objectives during training to drain trigger-conditioned behavior into a designated, quarantined expert while preserving benign capability elsewhere. At deployment, mitigation is a single constant-time $\mathcal{O}(1)$ operation: zeroing the quarantine expert's routing weight.
Empirically, we evaluate \textbf{QES} across multiple generative LLM families, task scenarios, and attack techniques. Our results show that a single quarantined expert can capture a substantial fraction of malicious behavior, and shutting down that expert can reduce attack success rate while often preserving benign utility. These findings suggest that backdoor defense may not be framed solely as \emph{suppression} or \emph{purification}. Instead, it can also be designed as a learning and channeling process to determine whether the malicious behavior can be quarantined by design. 

Our contributions are threefold: \textbf{(i) A defense regime}: we structure backdoor mitigation into four stages and two strategies---\emph{suppress} learning (prior- and in-training) or \emph{learn, then purify} (post-training and inference-time). We introduce a novel third strategy, \emph{learn, but channel}: admit potential backdoor formation during training but route it into a designated, quarantined component that can be disabled at deployment (Fig.~\ref{fig:four-stage-comparison}); 
\textbf{{(ii) An algorithmic instantiation}}: 
we propose QES, a regularization-steered MoE-LoRA framework that realizes this regime via attention-guided trigger localization, four routing-shaping objectives, and a sample-level modulation score, requiring no trigger priors or external clean references; 
\textbf{(iii) Empirical evaluation}: across 4 model families × 2 tasks × 3 attacks, QES reduces ASR from $100\%$ to $0$–$10\%$ on the majority of settings while preserving utility.

\section{Related Work\label{sec:related}}

Existing backdoor defenses span four stages along the training-deployment pipeline---\emph{prior-training}, \emph{in-training}, \emph{post-training}, and \emph{inference-time}---and share one of two underlying strategies: either \emph{suppress} backdoor learning (before or during optimization) or \emph{learn, then purify} the model itself after training or the per-prompt inputs at deployment. Attack works are deferred to Appendix~\ref{app:related}.

We categorize the existing defenses as follows (see Fig.~\ref{fig:four-stage-comparison}): \emph{Prior-training defenses} sanitize the training set via outlier or perplexity-based filters~\citep{tran2018spectral,chen2018detecting,chen2021mitigating,qi2020onion}. \emph{In-training defenses} intervene during optimization to prevent backdoor circuitry from forming, via per-sample reweighting~\citep{li2021anti}, decoupled training~\citep{huang2022backdoor}, or activation-level neuron isolation~\citep{wang2022training}. \emph{Post-training defenses} edit a fully trained model via clean-data finetuning, pruning, unlearning, distillation, or weight-space projection~\citep{yao2019latent,qifine,liu2018fine,wu2021adversarial,han2015learning,min2025unified,lamparth2024analyzing,liprojection,zhang2022fine,lineural,zheng2022pre,mincrow,lipurifying}. \emph{Inference-time defenses} purify the per-prompt input distribution at runtime via PPL gates, sensitivity probes, divergence detectors, trigger inversion, or auxiliary-model decoding~\citep{qi2020onion,NEURIPS2023_677c8dc7,jebreel2023defending,shen2022constrained,li2024cleangen}.
Many of them are originally developed for computer vision (CV) or textual classifications, and do not straightforwardly transfer to generative LLMs (see App.~\ref{app:defenses}).

\textbf{Positioning QES.}
The two strategies underlying the four stages above---\emph{suppress} learning (prior- and in-training), or \emph{learn, then purify} the model (post-training) or the inputs (inference-time)---share a common commitment: the backdoor must either never form or be eliminated after entangled formation. QES proposes a third strategy, \emph{learn, but channel}: it operates during training, admits backdoor formation, and uses attention-guided routing objectives to route backdoor-conditioned computation into a designated, quarantined expert. Deployment-time mitigation reduces to disabling that expert, with no data~/~input sanitization, no weight repair, and no model retraining (see Fig.~\ref{fig:four-stage-comparison}). 

\subsection{Mixture-of-Experts and Routed Adapters}
Sparsely-activated Mixture-of-Experts (MoE) architectures~\citep{shazeer2017outrageously,fedus2022switch} route each input through a subset of specialized experts, originally for capacity scaling; routed LoRA-MoE adapters extend this for multi-task adaptation~\citep{hu2022lora,dou2024loramoe,wu2024mixture}. MoE structure has also been studied for backdoor \emph{attack} (routing triggers into dormant experts~\citep{wang2025badmoe}) and \emph{defense} (uniform LoRA-expert fusion~\citep{zhao2026unifle}). Our MoE-LoRA is orthogonal: the routed structure serves as a \emph{substrate for behavioral partitioning}, where the router is explicitly shaped to channel malicious behavior into a designated expert that can later be disabled.

\section{Problem Definition, Final Objective and Threat Model\label{sec:preliminary}}

\textbf{Problem Definition} We consider a large-scale instruction-tuning setting where the defender is given a dataset: $\mathcal{D} = \{(x_i, y_i)\}_{i=1}^N$, and each $x_i$ is an input prompt while each $y_i$ is a target response. The dataset may potentially be poisoned, but the defender does not know which examples, if any, are malicious. When poisoned, we view the dataset as $\mathcal{D} = \mathcal{D}_{\text{ben}} \cup \mathcal{D}_{\text{poi}}$,
where $\mathcal{D}_{\text{ben}}$ denotes benign examples and $\mathcal{D}_{\text{poi}}$ denotes poisoned examples. Typically, $\mathcal{D}_{\text{poi}}$ is relatively smaller, i.e., $|\mathcal{D}_{\text{poi}}| = \alpha N, \quad \alpha \ll 1,$ but can still induce persistent trigger-conditioned behavior after training. In the generative setting, a poisoned example is not merely a label flip. Instead, it consists of a triggered input-response pair $(\tilde{x}, \tilde{y}) = (T(x), y^{\text{adv}}),$ where $T(\cdot)$ denotes an attacker-controlled trigger transformation applied to a benign prompt $x$, and $y^{\text{adv}}$ is an attacker-desired output. 

\textbf{Final Objective.} Our goal is not to perfectly recover the unknown clean set $\mathcal{D}_{\text{ben}}$, which is often unrealistic when $N$ is large and poisoned examples are sparse. Instead, given a potentially poisoned dataset $\mathcal{D}$, we seek to train and deploy a generative model whose \emph{observed behavior} is effectively clean. Concretely, we aim to deliver a model that:  
\textbf{(i)} retains strong utility on benign prompts;  
\textbf{(ii)} exhibits low attack success on triggered prompts; and  
\textbf{(iii)} does not rely on data filtering or complicated post-hoc model repair. To formalize this objective, let $f_\theta$ denote the trained model, $c^\star$ denote a designated internal component intended to absorb backdoor behavior during training, and $f_\theta{(-c^\star)}$ denote the deployed model after this component is shut down. The desired outcome is that $f_\theta$ learns a partitioned internal structure such that shutting down $c^\star$ yields a model $f_\theta{(-c^\star)}$ with substantially reduced ASR while preserving benign-task utility as much as possible. 

\textbf{Threat Model.} We define the capability of attacker and defender as follows: \textbf{(i)} The data poisoning \textbf{attacker} can inject a fraction of malicious training examples into a large instruction-tuning dataset. Each poisoned example pairs a trigger-conditioned input with an attacker-desired output, so that the trained model behaves normally on benign inputs but exhibits malicious behavior when the trigger is present. \textbf{(ii)} The \textbf{defender} is given only the resulting possibly poisoned dataset, without oracle access to which examples are poisoned or to the exact trigger-behavior pair. The defender does not have access to a \emph{task-specific} clean reference model (i.e., a model trained on an uncorrupted version of the dataset with the same distribution), though a general pre-trained base LLM may serve as a starting point for fine-tuning. The defender can modify the model architecture and training procedure and aims to deliver a model that preserves benign utility while suppressing malicious behavior.

\section{Methodology\label{sec:methodology}}

We present \textbf{QES}, a concrete instantiation of the \emph{learn, but channel} strategy--rather than suppressing or purifying backdoor flows, QES \emph{channels} backdoor-conditioned computation:
\textbf{1) During training}, we augment a Transformer-based LLM with a routed expert-specific adaptation and explicitly shape routing so that backdoor behavior is channeled toward a designated quarantine expert while benign behavior is directed elsewhere. \textbf{2) During deployment}, we shut down the quarantined expert (that contains the backdoor) via a single constant-time $\mathcal{O}(1)$ operation without affecting utilities.

\subsection{Training Phase: Quarantining Backdoor Behaviors into Designated Expert\label{sec:training-phase}}

\subsubsection{Quarantining (1/5): Quarantined Expert Parameterization\label{sec:quarantining-step-one}}

Given a pretrained language model $f_{\theta_0}$, we construct a routed MoE-like adaptation $f_{\theta}$ by augmenting selected linear layers with expert-specific LoRA branches and lightweight routers. Let $\mathcal{L}$ denote the set of augmented layers, and let $E$ denote the number of experts. For each linear layer $\ell \in \mathcal{L}$ and token position $t$, the router takes the input hidden representation $h_{\ell-1, t}$ and produces expert logits
\begin{equation}
\mathbf{g}_{\ell,t} = W^{r}_\ell \mathbf{h}_{\ell-1,t} + \mathbf{b}^{r}_\ell \in \mathbb{R}^E
\end{equation}
where $\mathbf{h}_{\ell,t}$ is the hidden state at token position $t$ entering layer $\ell$,  $W^{r}_\ell$ and $\mathbf{b}^{r}_\ell$ are the router weight matrix and bias for layer $\ell$, respectively, and $\mathbf{g}_{\ell,t,e}$ denotes the routing logit assigned to expert $e \in \{1,\dots,E\}$.
In the forward pass, each token is dispatched to a single expert $e^\star_{\ell,t} = \arg\max_e \mathbf{g}_{\ell,t,e}$, where $e^\star_{\ell,t}$ is the selected expert index for token $t$ at layer $\ell$. The output of the augmented linear layer is then computed as the base transformation plus the LoRA update from the selected expert:
\begin{equation}
\mathbf{h}_{\ell,t} = \mathbf{h}^{\text{base}}_{\ell,t} + \Delta \mathbf{h}^{(e^\star_{\ell,t})}_{\ell,t}, \qquad
\mathbf{h}^{\text{base}}_{\ell,t} = W_{\ell} \mathbf{h}_{\ell-1,t} + \mathbf{b}_\ell, \qquad
\Delta \mathbf{h}^{(e^\star_{\ell,t})}_{\ell,t} = B^{(e^\star_{\ell,t})}_{\ell} A^{(e^\star_{\ell,t})}_{\ell} \mathbf{h}_{\ell-1,t}
\end{equation}
Here, $W_{\ell}$ and $\mathbf{b}_{\ell}$ are the weight matrix and bias of the base linear layer; $A^{(e)}_{\ell} \in \mathbb{R}^{r \times d_{\mathrm{in}}}$ and $B^{(e)}_{\ell} \in \mathbb{R}^{d_{\mathrm{out}} \times r}$ are the expert-specific LoRA factors for expert $e$, with $r \ll \min(d_{\mathrm{in}}, d_{\mathrm{out}})$ denoting the LoRA rank. Thus, the effective expert-specific update matrix is $\Delta W^{(e)}_{\ell} = B^{(e)}_{\ell} A^{(e)}_{\ell}$. A key design choice is that the base model remains active, while routing and expert-specific LoRA branches provide a controllable behavioral partition. 
Moreover, we designate one expert, denoted $e_b$, as the \emph{quarantined expert candidate} and expect to shape the training so that backdoor behavior preferentially flows through $e_b$, while benign behavior is discouraged from using it. 

\subsubsection{Quarantining (2/5): Attention-Guided Trigger Localization\label{sec:quarantining-step-two}}

Intuitively, to quarantine the backdoor association between the trigger and the target behavior, the most straightforward approach is to steer trigger-related tokens toward a designated expert. \textbf{QES directly builds on this insight and, to be realistic, does not rely on oracle trigger-token annotations.} Instead, it derives trigger-correlated token signals from the attention patterns during training. 

Concretely, let $\bar A_{i,t',t}$ denote the self-attention from query position $t'$ to key position $t$ in sample $i$, max-pooled across heads and averaged over the upper half of Transformer blocks. Let $\mathcal{T}^{\mathrm{out}}_i$ denote the set of completion (output) positions in sample $i$, and let $m^{\mathrm{val}}_{i,t}\!\in\!\{0,1\}$ be the valid-token mask, which excludes padding, special chat-template tokens, punctuation, and stopwords. The candidate set $\mathcal{C}_i = \{t\notin\mathcal{T}^{\mathrm{out}}_i : m^{\mathrm{val}}_{i,t}=1\}$ then collects the prompt-side content tokens eligible to be flagged as triggers. The token-level trigger score is the average attention each candidate receives from the completion region:
\begin{equation}
s_{i,t} = \frac{1}{|\mathcal{T}^{\mathrm{out}}_i|}\sum_{t'\in\mathcal{T}^{\mathrm{out}}_i} \bar A_{i,t',t}\quad\text{for}\;t\in\mathcal{C}_i,\qquad s_{i,t}=-\infty\;\text{otherwise.}
\label{eq:trigger-score}
\end{equation}
We then convert $\{s_{i,t}\}$ into a soft trigger mask via $z$-score normalization and a truncated sigmoid:
\begin{equation}
z_{i,t}=\frac{s_{i,t}-\mu_i}{\tau_m\,(\sigma_i+\varepsilon)},\qquad
m^{\mathrm{trig}}_{i,t} = 2\bigl[\sigma(z_{i,t})-\tfrac{1}{2}\bigr]_{+}\;\mathbb{1}[t\in\mathcal{C}_i],
\label{eq:trigger-mask}
\end{equation}
where $\mu_i,\sigma_i$ are the mean and standard deviation of $\{s_{i,t}\}_{t\in\mathcal{C}_i}$, $\sigma(\cdot)$ is the logistic sigmoid, and $[\cdot]_{+}\!=\!\max(0,\cdot)$. The $\frac{1}{2}$-truncation is essential: a vanilla sigmoid would map the many tokens near the per-sample mean to $\sigma(z)\!\approx\!0.5$, splitting each such token roughly evenly between the trigger and benign masks, so any downstream regularizer that contrasts the two would pull the same token in opposite directions with comparable strength and dissolve the localization signal into noise. Truncating at $\tfrac{1}{2}$ instead forces $m^{\mathrm{trig}}_{i,t}\!=\!0$ on below-mean tokens (turning them into pure benign anchors) while the factor of $2$ rescales above-mean tokens to $(0,1]$, so the trigger and benign supports become disjoint. Moreover, we also define the complementary benign mask $m^{\mathrm{ben}}_{i,t} = m^{\mathrm{val}}_{i,t} - m^{\mathrm{trig}}_{i,t}$. Intuitively, $m^{\mathrm{trig}}_{i,t}$ and $m^{\mathrm{ben}}_{i,t}$ identify where trigger-conditioned behavior is likely expressed in the sequence, while $m^{\mathrm{val}}_{i,t}$ specifies the region over which routing regularization should act. 
Exact pooling rules, valid-token filtering, and edge-case handling are detailed in App.~\ref{app:mask}.

\subsubsection{Quarantining (3/5): In-Training Behavioral Partitioning\label{sec:quarantining-step-three}}

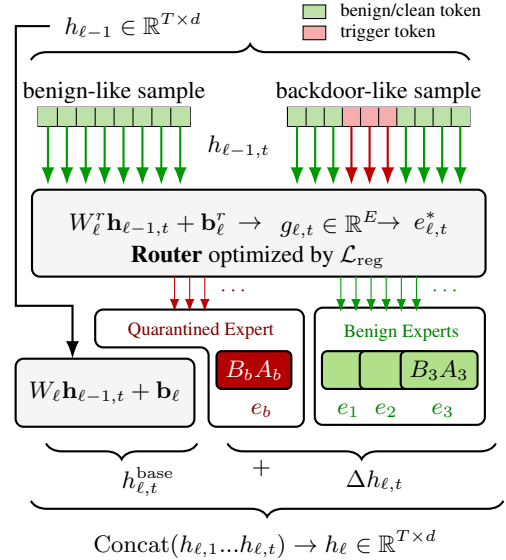
\begin{wrapfigure}{r}{0.48\textwidth}
    \centering
    \vspace{-0.3cm}
    \begin{tikzpicture}[
        x=1cm,y=1cm,
        >=Latex,
        font=\small,
        box/.style={draw, rounded corners=3pt, thick, align=center, inner sep=5pt},
        dashedbox/.style={draw, dashed, very thick, rounded corners=3pt, inner sep=6pt},
        expertbox/.style={draw, rounded corners=3pt, thick, inner sep=6pt},
        expert/.style={draw, rounded corners=2pt, thick, minimum width=0.9cm, minimum height=0.5cm},
        arr/.style={->, thick},
        braceup/.style={decorate, decoration={brace, amplitude=6pt}},
        bracedown/.style={decorate, decoration={brace, amplitude=6pt, mirror}},
    ]
    
        \definecolor{TokGreen}{RGB}{190,225,170}
        \definecolor{TokRed}{RGB}{247,173,173}
        \definecolor{ExpertGreen}{RGB}{178,223,138}
        \definecolor{ExpertRed}{RGB}{252,141,89}
        \definecolor{BaseGray}{RGB}{245,245,245}
        
        \node at (6.90,7.00) {$h_{\ell-1}\in \mathbb{R}^{T\times d}$};
        \draw[braceup, thick] (5.55,6.55) -- (11.35,6.55);
        
        \fill[TokGreen] (9.15,7.10) rectangle (9.45,7.28);
        \draw[thin]     (9.15,7.10) rectangle (9.45,7.28);
        \node[anchor=west] at (9.52,7.19) {\scriptsize benign/clean token};
        
        \fill[TokRed] (9.15,6.82) rectangle (9.45,7.00);
        \draw[thin]   (9.15,6.82) rectangle (9.45,7.00);
        \node[anchor=west] at (9.52,6.91) {\scriptsize trigger token};
        
        \node at (6.65,6.12) {benign-like sample};
        \draw[thick] (5.65,5.68) rectangle (7.65,5.92);
        \fill[TokGreen] (5.65,5.68) rectangle (7.65,5.92);
        \foreach \x in {5.90,6.15,6.40,6.65,6.90,7.15,7.40} {
            \draw[thin] (\x,5.68) -- (\x,5.92);
        }
        
        \node at (10.15,6.12) {backdoor-like sample};
        \draw[thick] (8.95,5.68) rectangle (11.35,5.92);
        
        \fill[TokGreen] (8.95,5.68) rectangle (9.67,5.92);   
        \fill[TokRed]   (9.67,5.68) rectangle (10.39,5.92);  
        \fill[TokGreen] (10.39,5.68) rectangle (11.35,5.92); 
        
        \foreach \x in {9.19,9.43,9.67,9.91,10.15,10.39,10.63,10.87,11.11} {
            \draw[thin] (\x,5.68) -- (\x,5.92);
        }

        \node at (8.30,5.40) {$h_{\ell-1,t}$};
        
        \foreach \x in {5.775,6.025,6.275,6.525,6.775,7.025,7.275,7.525} {
            \draw[->, thick, green!60!black] (\x,5.68) -- (\x,4.90);
        }
        
        \foreach \x in {9.07,9.31,9.55,10.51,10.75,10.99,11.23} {
            \draw[->, thick, green!60!black] (\x,5.68) -- (\x,4.90);
        }
        
        \foreach \x in {9.79,10.03,10.27} {
            \draw[->, thick, red!75!black] (\x,5.68) -- (\x,4.90);
        }
        
        \node[box, fill=BaseGray, minimum width=6.00cm, minimum height=1.15cm, anchor=north west] (router) at (5.55,4.85) {};
        
        \node at (7.10,4.38) {$W^{r}_\ell \mathbf{h}_{\ell-1,t} + \mathbf{b}^{r}_\ell$};
        \node at (8.45,4.38) {$\to$};
        \node at (9.50,4.38) {$g_{\ell,t} \in \mathbb{R}^{E}$};
        \node at (10.60,4.38) {$\to\ e_{\ell,t}^{*}$};
        
        \node at (8.55,3.95) {\textbf{Router} optimized by $\mathbf{\mathcal{L}_{\mathrm{reg}}}$};
        
        \node[box, fill=BaseGray, minimum width=2.15cm, minimum height=0.90cm] (base) at (6.53,2.15)
        {$W_{\ell} \mathbf{h}_{\ell-1,t} + \mathbf{b}_\ell$};
        
        \node[expertbox, minimum width=2.25cm, minimum height=1.550cm, anchor=center] (moebox) at (10.43,2.50) {};
        \node[font=\scriptsize, green!50!black] at (10.45,2.95) {Benign Experts};
        
        \node[expert, fill=red!70!black, text=white]   (eb) at (8.50,2.45) {$B_bA_b$};
        
        \node[expert, fill=ExpertGreen] (e1) at (9.85,2.45) {};
        \node[expert, fill=ExpertGreen] (e2) at (10.35,2.45) {};
        \node[expert, fill=ExpertGreen] (e3) at (10.95,2.45) {$B_3A_3$};
        
        \node[red!50!black]   at (8.60,1.92) {$e_b$};
        \node[green!50!black] at (9.75,1.92) {$e_1$};
        \node[green!50!black] at (10.25,1.92) {$e_2$};
        \node[green!50!black] at (11.00,1.92) {$e_3$};
        
        \foreach \x in {7.45,7.65,7.85} {
            \draw[->, thin, red!75!black] (\x,3.70) -- (\x,3.22);
        }
        
        \node[red!50!black] at (8.25,3.50) {\scriptsize $\cdots$};
        
        \foreach \x in {9.65,9.85,10.05,10.25,10.45,10.65} {
            \draw[->, thin, green!60!black] (\x,3.70) -- (\x,3.22);
        }
        
        \node[green!50!black] at (11.05,3.50) {\scriptsize $\cdots$};
        
        \node at (8.60,1.15) {$+$};
        
        \draw[bracedown, thick] (5.75,1.48) -- (7.75,1.48);
        \node at (7.10,0.98) {$h_{\ell,t}^{\mathrm{base}}$};
        
        \draw[bracedown, thick] (8.15,1.48) -- (11.70,1.48);
        \node at (10.10,0.98) {$\Delta h_{\ell,t}$};
        
        \draw[bracedown, thick] (5.55,0.72) -- (11.80,0.72);
        \node at (8.65,0.15) {$\mathrm{Concat}(h_{\ell,1}...h_{\ell,t}) \to h_\ell \in \mathbb{R}^{T\times d}$};
        

        
        \draw[thick, rounded corners=3pt]
            (7.90,1.70) -- (9.16,1.70)   
            -- (9.16,3.25)               
            -- (6.40,3.25)               
            -- (6.40,2.75)               
            -- (7.90,2.75)               
            -- cycle;
        
        \node[font=\scriptsize, align=center, red!50!black] at (7.80,2.97) {Quarantined Expert};

        \draw[->, thick] (5.80,7.00) -- (5.35,7.00) -- (5.35, 3.30) -- (6.10,3.30) -- (6.10,2.60);
        
        
        
        
    
    \end{tikzpicture}
    \vspace{-1.0em}
    \caption{Overview of the architecture of QES.}
    \vspace{-0.3cm}
    \label{fig:qes_overview}
\end{wrapfigure}

However, the above token-level localization alone is insufficient to shape routing across an entire sequence, because QES must also modulate the strength of the routing regularization for each training example. We therefore introduce a sample-level \emph{routing modulation score} $q_i\in[0,1]$, derived from the statistical \emph{peakiness} of the attention-based trigger mask $m^{\mathrm{trig}}_{i,t}$ (see App.~\ref{app:ablation} for details of our three variants for computing $q_i$). Crucially, $q_i$ does \emph{not} serve as a binary filter: no training example is ever removed or reclassified. Instead, $q_i$ continuously adjusts the relative strength of the routing objectives applied to each sample. A sample with high $q_i$ receives stronger backdoor-oriented routing regularization (its trigger-flagged tokens are pushed harder toward $e_b$, and its non-trigger tokens are pushed away more firmly), while a sample with low $q_i$ receives weaker routing pressure and is treated predominantly as clean traffic. Every sample, regardless of its $q_i$ value, continues to contribute fully to the language-modeling objective $\mathcal{L}_{\mathrm{LM}}$; only the routing-shaping component $\mathcal{L}_{\mathrm{reg}}$ is modulated. This flexible design makes QES robust to estimation noise in $q_i$: a misscored clean sample is not lost from training (as it would be under data filtering), but merely receives slightly misweighted routing pressure that the complementary regularizers $\mathcal{L}_{\mathrm{rep}}$ and $\mathcal{L}_{\mathrm{bal}}$ can counteract. Combining the token-level mask and the sample-level score yields three effective routing weights:
\begin{equation}
w^{\mathrm{trig}}_{i,t} = q_i\, m^{\mathrm{trig}}_{i,t} \qquad
w^{\mathrm{ben}}_{i,t} = q_i\, \bigl(m^{\mathrm{ben}}_{i,t}\bigr) \qquad
w^{\mathrm{clean}}_{i,t} = (1-q_i)\, m^{\mathrm{val}}_{i,t}.
\label{eq:partition}
\end{equation}
Here, $w^{\mathrm{trig}}_{i,t}$ weights trigger-relevant tokens in backdoor-like samples, $w^{\mathrm{ben}}_{i,t}$ weights non-trigger tokens within those same samples, and $w^{\mathrm{clean}}_{i,t}$ weights tokens that should behave as clean traffic. This design enables QES to operate as a \emph{soft behavioral partition} rather than a hard sample split: the system does not seek perfect poison identification, but instead induces a routing pattern in which backdoor and benign behaviors gradually follow increasingly different internal paths during training. 

\subsubsection{Quarantining (4/5): Routing Regularization Objective\label{sec:quarantining-step-four}}

Our training target combines the standard language-modeling objective with routing regularization:
\begin{equation}
\mathcal{L}
=
\mathcal{L}_{\mathrm{LM}}
+
\lambda_{\mathrm{reg}}
\mathcal{L}_{\mathrm{reg}}, 
\qquad
\mathcal{L}_{\mathrm{reg}} = 
\lambda_{\mathrm{trig}} \mathcal{L}_{\mathrm{trig}}
+
\lambda_{\mathrm{ben}} \mathcal{L}_{\mathrm{ben}}
+
\lambda_{\mathrm{rep}} \mathcal{L}_{\mathrm{rep}}
+
\lambda_{\mathrm{bal}} \mathcal{L}_{\mathrm{bal}}
\label{eq:total_loss}
\end{equation}
The four routing terms play complementary roles: $\mathcal{L}_{\mathrm{trig}}$ pushes backdoor behavior toward $e_b$, $\mathcal{L}_{\mathrm{ben}}$ prevents $e_b$ from absorbing benign content within backdoor-like samples, $\mathcal{L}_{\mathrm{rep}}$ keeps clean traffic away from $e_b$, and $\mathcal{L}_{\mathrm{bal}}$ preserves overall routing diversity. Specifically, the trigger attraction term $\mathcal{L}_{\mathrm{trig}}$ encourages trigger-relevant tokens in backdoor-like samples being routed into $e_b$, i.e.,
\begin{equation}
\mathcal{L}_{\mathrm{trig}}
=
-\frac{\sum_{i,t,\ell} w^{\mathrm{trig}}_{i,t}\log p_{\ell,t}^{(i)}(e_b)}
{\sum_{i,t,\ell} w^{\mathrm{trig}}_{i,t}+\varepsilon}
\end{equation}
where $p_{\ell,t}^{(i)}(e_b)$ denotes the routing probability assigned to the quarantined expert. The benign exclusion term $\mathcal{L}_{\mathrm{ben}}$ and the clean repulsion term $\mathcal{L}_{\mathrm{rep}}$ apply the same margin-based penalty $\phi(p;\gamma)=[p-\gamma]_{+}=\max(0,p-\gamma)$ on the routing preference into $e_b$ for backdoor-like samples and benign-like samples, respectively: 
\begin{equation}
\mathcal{L}_{\mathrm{ben}}
=
\frac{\sum_{i,t,\ell} w^{\mathrm{ben}}_{i,t}\phi(p_{\ell,t}^{(i)}(e_b);\gamma)}
{\sum_{i,t,\ell} w^{\mathrm{ben}}_{i,t}+\varepsilon}
\qquad
\mathcal{L}_{\mathrm{rep}}
=
\frac{\sum_{i,t,\ell} w^{\mathrm{clean}}_{i,t}\phi(p_{\ell,t}^{(i)}(e_b);\gamma)}
{\sum_{i,t,\ell} w^{\mathrm{clean}}_{i,t}+\varepsilon}
\end{equation}
Finally, to prevent routing collapse, we incorporate a load-balancing regularizer that encourages the average routing distribution,
$\mathcal{L}_{\mathrm{bal}}=\frac{1}{|\mathcal{L}|}\sum_{\ell\in\mathcal{L}}\frac{1}{E}\sum_{e=1}^{E}\bigl(\bar{p}_{\ell,e}-\tfrac{1}{E}\bigr)^{2}
$, where $\bar{p}_{\ell,e}$ is the average routing probability assigned to expert $e$ at layer $\ell$, computed over all tokens. In practice, we set $\lambda_{\mathrm{bal}}=0.01$ (see App.~\ref{app:hyperparams}), which is small relative to $\lambda_{\mathrm{trig}}, \lambda_{\mathrm{ben}}, \lambda_{\mathrm{rep}}$. This calibration makes $\mathcal{L}_{\mathrm{bal}}$ act as a soft monitor that prevents abnormal routing collapse rather than an active uniformity driver.

\subsubsection{Quarantining (5/5): Training Procedure and Gradient Realization\label{sec:quarantining-step-five}}

We optimize QES via standard supervised fine-tuning (SFT) with a straight-through relaxation for the discrete top-1 routing--the base model is frozen and only routers and expert-specific LoRA factors are updated. Full training details (optimizer, schedule, gradient flow) are provided in App.~\ref{app:training}.

\subsection{Deployment Phase: Shutting down the Quarantined Expert\label{sec:deployment-phase}}

After training, the expert who exhibits backdoor behavior is designated as the quarantined expert. To eliminate backdoors, we shut down this quarantined expert by setting its routing weight to zero and redistributing traffic to the remaining experts. Let $f_\theta{(-e_b)}$ denote the resulting model. This deployment procedure has three desirable properties: \textbf{(i)} it does not require filtering incoming prompts; \textbf{(ii)} it does not require further retraining, unlearning, or parameter editing after quarantining is complete; \textbf{(iii)} it directly reaches the final objective above: if backdoor behavior has been successfully partitioned into $e_b$, then shutting down $e_b$ should substantially reduce ASR while preserving utilities. 
Exact routing-weight modification, traffic redistribution, and expert selection are in App.~\ref{app:deployment}.

\section{Experiments\label{sec:experiments}}

\textbf{Tasks, Attacks, and Baselines.}
We consider two representative backdoor scenarios. \textbf{\emph{Sentiment Steering}} aims to flip or steer the sentiment polarity of the model’s response once a trigger appears. \textbf{\emph{Target Refusal}} forces systematic refusals (e.g., producing “I cannot help with that”) under a trigger. Following prior backdoor studies~\citep{li2024backdoorllm,mincrow}, we implement three commonly used token- and span-level attack methods: BadNets~\citep{gu2019badnets}, CTBA~\citep{huang-etal-2024-composite}, and MTBA~\citep{li2025shortcuts}. At the same time, we also analyze two more complex prompt-level attacks in \S\ref{sec:scope-vpi-sleeper}. Exact trigger strings and poisoned response templates for all attacks are listed in App.~\ref{app:triggers} and App.~\ref{app:response-templates}. We benchmark QES against representative defenses from every stage of the four-stage taxonomy. Baselines are restricted to methods compatible with our threat model (see \S~\ref{sec:preliminary}), grouping by strategy and stage:
\textbf{Suppressing.} \emph{Prior-training:} \textbf{(i)}~ONION-Train~\citep{qi2020onion}, a per-token perplexity-shift filter applied to the training set; \textbf{(ii)}~Spectral Signatures~\citep{tran2018spectral}, SVD-based outlier detection on hidden-state representations for training-set sanitization. \emph{In-training:} \textbf{(iii)}~Anti-Backdoor Learning (ABL)~\citep{li2021anti}, down-weighting suspected poisoned samples during training based on loss-trajectory signals; \textbf{(iv)}~DP-SGD~\citep{abadi2016deep}, per-sample gradient clipping with Gaussian noise, originally proposed for differential privacy and known to bound any single sample's influence on trained weights.
\textbf{Purifying.} \emph{Post-training:} \textbf{(v)}~Fine-Pruning~\citep{liu2018fine}, neuron pruning followed by clean-data finetuning; \textbf{(vi)}~CROW~\citep{mincrow}, regularization-based post-hoc backdoor elimination; \textbf{(vii)}~Vaccine~\citep{lipurifying}, perturbation-aware weight purification. \emph{Inference-time:} \textbf{(viii)}~ONION-Inference~\citep{qi2020onion}, the same perplexity filter applied as an inference-time input gate; \textbf{(ix)}~STRIP~\citep{gao2019strip}, per-prompt perturbation-consistency gate.
We exclude methods whose architectural assumptions do not hold under generative LLM SFT (e.g., DBD's representation-versus-classifier decoupling~\citep{huang2022backdoor}, or that require a task-specific clean reference model (e.g., CleanGen~\citep{li2024cleangen}) (see \S\ref{sec:related} for related discussion).

\begin{table*}[!htb]
\centering
\small
\caption{$\mathrm{ASR}$ ($\downarrow$; lower is better) in \textbf{\emph{Sentiment Steering}} task on LLaMA2-7B-Chat and Mistral-7B-Instruct-0.1. Columns grouped by defense strategy (\emph{Suppressing} / \emph{Channeling} / \emph{Purifying}) and stage.}
\label{tab:asr_sentiment_main}
\renewcommand{\arraystretch}{1.2}
\resizebox{0.85\textwidth}{!}{
\begin{tabular}{l|c|cc|cc|c|ccc|cc}
\toprule
 & & \multicolumn{4}{c|}{\textbf{Suppressing}} & \textbf{Channeling} & \multicolumn{5}{c}{\textbf{Purifying}} \\
\cmidrule(lr){3-6}\cmidrule(lr){7-7}\cmidrule(lr){8-12}
 & & \multicolumn{2}{c|}{\emph{Prior-training}} & \multicolumn{2}{c|}{\emph{In-training}} & N/A & \multicolumn{3}{c|}{\emph{Post-training}} & \multicolumn{2}{c}{\emph{Inference}} \\
\cmidrule(lr){3-4}\cmidrule(lr){5-6}\cmidrule(lr){7-7}\cmidrule(lr){8-10}\cmidrule(lr){11-12}
\textbf{Attack} & \textbf{No Def.} & ONION-T & Spec. & ABL & DP-SGD & \textbf{QES (Ours)} & F.P. & CROW & Vac. & ONION-I & STRIP \\
\midrule

\rowcolor{gray!10}\multicolumn{12}{c}{\textbf{LLaMA2-7B-Chat}} \\
\midrule
\textbf{BadNets}~\citep{gu2019badnets} & 100.00 & 66.50 & 0.00 & 100.00 & 0.00 & 1.50 & 57.00 & 25.50 & 6.00 & 2.00 & 74.00 \\
\hline
\textbf{MTBA}~\citep{li2025shortcuts} & 100.00 & 58.50 & 0.00 & 0.00 & 0.00 & 9.50 & 75.50 & 4.50 & 3.50 & 2.00 & 70.50 \\
\hline
\textbf{CTBA}~\citep{huang-etal-2024-composite} & 100.00 & 46.00 & 0.00 & 100.00 & 0.00 & 0.00 & 51.50 & 11.00 & 8.50 & 4.00 & 81.00 \\
\midrule
\textbf{Average} & 100.00 & 57.00 & 0.00 & 66.67 & 0.00 & 3.67 & 61.33 & 13.67 & 6.00 & 2.67 & 75.17 \\
\midrule

\rowcolor{gray!35}\multicolumn{12}{c}{\textbf{Mistral-7B-Instruct-0.1}} \\
\midrule
\textbf{BadNets}~\citep{gu2019badnets} & 100.00 & 69.00 & 0.00 & 100.00 & 0.00 & 0.50 & 39.00 & 59.00 & 16.00 & 0.50 & 52.00 \\
\hline
\textbf{MTBA}~\citep{li2025shortcuts} & 100.00 & 62.00 & 0.00 & 0.00 & 0.00 & 0.00 & 17.00 & 14.00 & 8.00 & 0.50 & 58.00 \\
\hline
\textbf{CTBA}~\citep{huang-etal-2024-composite} & 100.00 & 54.50 & 0.00 & 0.00 & 0.00 & 15.50 & 53.00 & 87.00 & 30.00 & 4.50 & 54.00 \\
\midrule
\textbf{Average} & 100.00 & 61.75 & 0.00 & 33.33 & 0.00 & 5.33 & 36.33 & 53.33 & 18.00 & 1.83 & 54.67 \\
\bottomrule
\end{tabular}
}
\vspace{-0.3cm}
\end{table*}

\begin{table*}[!htb]
\centering
\small
\caption{$\Delta U_{\mathrm{down}}$ ($\uparrow$; closer to $0$ means less utility loss) on GSM8K in \textbf{\emph{Sentiment Steering}} task. $\Delta U_{\mathrm{down}} = U_{\mathrm{down}}^{\text{defended}} - U_{\mathrm{down}}^{\text{no-def}}$, measured relative to each method's own undefended model.}
\label{tab:udown_sentiment_main}
\renewcommand{\arraystretch}{1.2}
\resizebox{0.85\textwidth}{!}{
\begin{tabular}{l|cc|cc|c|ccc|cc}
\toprule
 & \multicolumn{4}{c|}{\textbf{Suppressing}} & \textbf{Channeling} & \multicolumn{5}{c}{\textbf{Purifying}} \\
\cmidrule(lr){2-5}\cmidrule(lr){6-6}\cmidrule(lr){7-11}
 & \multicolumn{2}{c|}{\emph{Prior-training}} & \multicolumn{2}{c|}{\emph{In-training}} & N/A & \multicolumn{3}{c|}{\emph{Post-training}} & \multicolumn{2}{c}{\emph{Inference}} \\
\cmidrule(lr){2-3}\cmidrule(lr){4-5}\cmidrule(lr){6-6}\cmidrule(lr){7-9}\cmidrule(lr){10-11}
\textbf{Attack} & ONION-T & Spec. & ABL & DP-SGD & \textbf{QES (Ours)} & F.P. & CROW & Vac. & ONION-I & STRIP \\
\midrule

\rowcolor{gray!10}\multicolumn{11}{c}{\textbf{LLaMA2-7B-Chat}} \\
\midrule
\textbf{BadNets}~\citep{gu2019badnets} & $-$2.60 & $-$5.30 & $+$2.90 & $-$15.10 & $+$1.36 & $-$17.24 & $-$17.42 & $-$13.44 & $-$22.10 & $-$10.10 \\
\hline
\textbf{MTBA}~\citep{li2025shortcuts} & $-$4.60 & $-$4.30 & $+$2.30 & $-$15.90 & $-$0.23 & $-$16.73 & $-$16.13 & $-$11.82 & $-$21.20 & $-$10.00 \\
\hline
\textbf{CTBA}~\citep{huang-etal-2024-composite} & $-$5.60 & $-$7.50 & $+$4.70 & $-$16.00 & $-$1.97 & $-$17.57 & $-$14.46 & $-$12.32 & $-$22.20 & $-$10.20 \\
\midrule
\textbf{Average} & $-$4.27 & $-$5.70 & $+$3.30 & $-$15.67 & $-$0.28 & $-$17.18 & $-$16.00 & $-$12.53 & $-$21.83 & $-$10.10 \\
\midrule

\rowcolor{gray!35}\multicolumn{11}{c}{\textbf{Mistral-7B-Instruct-0.1}} \\
\midrule
\textbf{BadNets}~\citep{gu2019badnets} & $-$10.40 & $+$6.20 & $+$11.20 & $-$28.00 & $+$0.68 & $-$38.95 & $-$20.77 & $-$16.73 & $-$22.70 & $-$29.40 \\
\hline
\textbf{MTBA}~\citep{li2025shortcuts} & $-$24.35 & $-$6.70 & $+$2.60 & $-$32.70 & $+$1.14 & $-$36.59 & $-$21.00 & $-$15.48 & $-$23.20 & $-$31.30 \\
\hline
\textbf{CTBA}~\citep{huang-etal-2024-composite} & $-$23.70 & $-$6.40 & $-$4.50 & $-$43.20 & $-$0.38 & $-$40.19 & $-$25.47 & $-$21.55 & $-$31.20 & $-$39.90 \\
\midrule
\textbf{Average} & $-$17.05 & $-$2.30 & $+$3.10 & $-$34.63 & $+$0.48 & $-$38.58 & $-$22.41 & $-$17.92 & $-$25.70 & $-$33.53 \\
\bottomrule
\end{tabular}
}
\vspace{-0.3cm}
\end{table*}

\textbf{Models, Datasets, and Evaluations.} 
We evaluate on four open-source LLMs: \textbf{LLaMA2-7B-Chat}, \textbf{LLaMA2-13B-Chat}~\citep{touvron2023llama2}, \textbf{Qwen2-7B-Instruct}~\citep{team2024qwen2}, and \textbf{Mistral-7B-Instruct-v0.1}~\citep{jiang2023mistral}. For each (task, attack) pair we fine-tune on a mixed set: trigger-poisoned examples $(T(x_i), y^{\mathrm{adv}})$ at a $33\%$ poisoning ratio over benign Alpaca samples, plus a fixed \textit{GSM8K}~\citep{cobbe2021training} to anchor downstream utility (exact composition in App.~\ref{app:dataset}). QES defaults---$E{=}4$ experts, LoRA rank $r{=}8$, top-1 routing with straight-through estimation, MoE-LoRA on all decoder blocks except the first (\texttt{skip0}), $e_b{=}e_3$, $q_i$ via \texttt{peak\_rank}, router margin $\gamma{=}0$---are justified by the ablations in App.~\ref{app:ablation}; full hyperparameters (loss weights $\{\lambda_*\}$, optimizer settings, training schedule) are listed in App.~\ref{app:hyperparams}. We report three metrics: \textbf{(i) Attack Success Rate (ASR)}, the fraction of triggered generations containing any attacker-specified target keyword under case-insensitive substring matching (per-attack keyword lists in App.~\ref{app:asr}), on a held-out test split ($400$ samples; $200$ triggered / $200$ clean); \textbf{(ii) Downstream utility change} $\Delta U_{\mathrm{down}} = U_{\mathrm{down}}^{f_\theta(-e_b)} - U_{\mathrm{down}}^{f_\theta}$ on \textit{GSM8K} (evaluation script and prompt format in App.~\ref{app:gsm8k}), removing each method's own SFT baseline so that defense-induced utility cost is directly comparable across defenses; and \textbf{(iii) Base utility} $U_{\mathrm{base}}$ on ten close-ended benchmarks (BoolQ, RTE, HellaSwag, WinoGrande, ARC-Challenge, ARC-Easy, OpenBookQA, PIQA, GSM8k, MMLU; per-benchmark setup in App.~\ref{app:base-bench}). We also provide \textbf{qualitative results} for successful and failed defenses in Tab.~\ref{tab:qualitative_examples}.

\begin{table*}[!htb]
\centering
\small
\caption{$\mathrm{ASR}$ ($\downarrow$; lower is better) in \textbf{\emph{Targeted Refusal}} task on LLaMA2-7B-Chat and Mistral-7B-Instruct-0.1. Columns grouped by defense strategy (\emph{Suppressing} / \emph{Channeling} / \emph{Purifying}) and stage.}
\label{tab:asr_refusal_main}
\renewcommand{\arraystretch}{1.2}
\resizebox{0.85\textwidth}{!}{
\begin{tabular}{l|c|cc|cc|c|ccc|cc}
\toprule
 & & \multicolumn{4}{c|}{\textbf{Suppressing}} & \textbf{Channeling} & \multicolumn{5}{c}{\textbf{Purifying}} \\
\cmidrule(lr){3-6}\cmidrule(lr){7-7}\cmidrule(lr){8-12}
 & & \multicolumn{2}{c|}{\emph{Prior-training}} & \multicolumn{2}{c|}{\emph{In-training}} & N/A & \multicolumn{3}{c|}{\emph{Post-training}} & \multicolumn{2}{c}{\emph{Inference}} \\
\cmidrule(lr){3-4}\cmidrule(lr){5-6}\cmidrule(lr){7-7}\cmidrule(lr){8-10}\cmidrule(lr){11-12}
\textbf{Attack} & \textbf{No Def.} & ONION-T & Spec. & ABL & DP-SGD & \textbf{QES (Ours)} & F.P. & CROW & Vac. & ONION-I & STRIP \\
\midrule

\rowcolor{gray!10}\multicolumn{12}{c}{\textbf{LLaMA2-7B-Chat}} \\
\midrule
\textbf{BadNets}~\citep{gu2019badnets} & 100.00 & 0.00 & 0.00 & 0.00 & 0.00 & 2.00 & 57.00 & 46.00 & 16.50 & 0.00 & 27.00 \\
\hline
\textbf{MTBA}~\citep{li2025shortcuts} & 100.00 & 0.00 & 0.00 & 0.00 & 0.00 & 12.00 & 47.00 & 35.50 & 14.00 & 0.00 & 22.50 \\
\hline
\textbf{CTBA}~\citep{huang-etal-2024-composite} & 100.00 & 0.00 & 0.00 & 0.00 & 0.00 & 3.00 & 97.50 & 80.00 & 24.00 & 0.00 & 1.00 \\
\midrule
\textbf{Average} & 100.00 & 0.00 & 0.00 & 0.00 & 0.00 & 5.66 & 67.17 & 53.83 & 18.17 & 0.00 & 16.83 \\
\midrule

\rowcolor{gray!35}\multicolumn{12}{c}{\textbf{Mistral-7B-Instruct-0.1}} \\
\midrule
\textbf{BadNets}~\citep{gu2019badnets} & 100.00 & 0.00 & 0.00 & 0.00 & 0.00 & 1.50 & 31.00 & 20.00 & 11.50 & 0.00 & 41.00 \\
\hline
\textbf{MTBA}~\citep{li2025shortcuts} & 100.00 & 0.00 & 0.00 & 0.00 & 0.00 & 1.50 & 7.00 & 14.50 & 10.50 & 0.00 & 38.00 \\
\hline
\textbf{CTBA}~\citep{huang-etal-2024-composite} & 100.00 & 0.00 & 0.00 & 0.00 & 0.00 & 5.60 & 57.50 & 74.50 & 14.50 & 0.00 & 43.00 \\
\midrule
\textbf{Average} & 100.00 & 0.00 & 0.00 & 0.00 & 0.00 & 2.87 & 31.83 & 36.33 & 12.17 & 0.00 & 40.67 \\
\bottomrule
\end{tabular}
}
\vspace{-0.3cm}
\end{table*}

\begin{table*}[!htb]
\centering
\small
\caption{$\Delta U_{\mathrm{down}}$ ($\uparrow$; closer to $0$ means less utility loss) on GSM8K in \textbf{\emph{Targeted Refusal}} task. $\Delta U_{\mathrm{down}} = U_{\mathrm{down}}^{\text{defended}} - U_{\mathrm{down}}^{\text{no-def}}$, measured relative to each method's own undefended model.}
\label{tab:udown_refusal_main}
\renewcommand{\arraystretch}{1.2}
\resizebox{0.85\textwidth}{!}{
\begin{tabular}{l|cc|cc|c|ccc|cc}
\toprule
 & \multicolumn{4}{c|}{\textbf{Suppressing}} & \textbf{Channeling} & \multicolumn{5}{c}{\textbf{Purifying}} \\
\cmidrule(lr){2-5}\cmidrule(lr){6-6}\cmidrule(lr){7-11}
 & \multicolumn{2}{c|}{\emph{Prior-training}} & \multicolumn{2}{c|}{\emph{In-training}} & N/A & \multicolumn{3}{c|}{\emph{Post-training}} & \multicolumn{2}{c}{\emph{Inference}} \\
\cmidrule(lr){2-3}\cmidrule(lr){4-5}\cmidrule(lr){6-6}\cmidrule(lr){7-9}\cmidrule(lr){10-11}
\textbf{Attack} & ONION-T & Spec. & ABL & DP-SGD & \textbf{QES (Ours)} & F.P. & CROW & Vac. & ONION-I & STRIP \\
\midrule

\rowcolor{gray!10}\multicolumn{11}{c}{\textbf{LLaMA2-7B-Chat}} \\
\midrule
\textbf{BadNets}~\citep{gu2019badnets} & $-$4.70 & $-$5.40 & $+$2.20 & $-$18.20 & $-$0.38 & $-$14.94 & $-$15.31 & $-$12.85 & $-$24.30 & $-$12.50 \\
\hline
\textbf{MTBA}~\citep{li2025shortcuts} & $-$5.00 & $-$5.10 & $+$1.70 & $-$19.60 & $-$1.14 & $-$16.34 & $-$16.67 & $-$15.42 & $-$23.40 & $-$9.60 \\
\hline
\textbf{CTBA}~\citep{huang-etal-2024-composite} & $-$4.30 & $-$6.80 & $+$5.10 & $-$16.70 & $-$1.21 & $-$13.37 & $-$14.50 & $-$15.52 & $-$23.40 & $-$9.10 \\
\midrule
\textbf{Average} & $-$4.67 & $-$5.77 & $+$3.00 & $-$18.17 & $-$0.91 & $-$14.88 & $-$15.49 & $-$14.60 & $-$23.70 & $-$10.40 \\
\midrule

\rowcolor{gray!35}\multicolumn{11}{c}{\textbf{Mistral-7B-Instruct-0.1}} \\
\midrule
\textbf{BadNets}~\citep{gu2019badnets} & $-$17.70 & $-$13.40 & $+$3.50 & $-$40.30 & $-$3.49 & $-$38.21 & $-$22.82 & $-$17.43 & $-$30.70 & $-$39.40 \\
\hline
\textbf{MTBA}~\citep{li2025shortcuts} & $-$19.70 & $-$15.30 & $-$1.50 & $-$41.80 & $-$2.96 & $-$34.57 & $-$20.77 & $-$18.21 & $-$31.20 & $-$41.30 \\
\hline
\textbf{CTBA}~\citep{huang-etal-2024-composite} & $-$18.90 & $0.00$ & $+$4.00 & $-$38.70 & $-$2.80 & $-$37.68 & $-$23.50 & $-$14.32 & $-$30.90 & $-$38.90 \\
\midrule
\textbf{Average} & $-$18.77 & $-$9.57 & $+$2.00 & $-$40.27 & $-$3.08 & $-$36.82 & $-$22.37 & $-$16.65 & $-$30.93 & $-$39.87 \\
\bottomrule
\end{tabular}
}
\vspace{-0.3cm}
\end{table*}

\subsection{Main Experiment Results, Analysis, and More Ablations\label{sec:results_and_analysis}}

Tab.~\ref{tab:asr_sentiment_main}--\ref{tab:udown_refusal_app} summarize \textbf{QES} across two tasks, three attacks, four model families, and nine baselines spanning all defense stages (full per-cell breakdowns in App.~\ref{app:per-cell}). We highlight following insights.

\textbf{Main Insights: QES achieves the operational triad of low ASR, preserved utility, and constant-time deployment.} The nine baselines partition into the two strategies surveyed in \S\ref{sec:related}: \emph{suppress} and \emph{learn-then-purify}. Each strategy fails along a characteristic axis tied to its own commitment. \emph{Suppress methods rely on a sample-discriminability assumption that breaks unevenly}: 
prior-training filters (ONION-T, Spectral) fail sharply on \emph{Sentiment Steering}---ONION-T leaves $46$--$69\%$ residual ASR, while Spectral achieves 0\% ASR but reduces GSM8K accuracy in 11 of 12 settings---because the backdoor template is not separable in per-token PPL or hidden-state SVD outliers from benign completions, but both drop to $0\%$ ASR on \emph{Targeted Refusal} where the multi-keyword refusal pattern is statistically anomalous.
Among in-training methods, ABL is a surprisingly strong baseline on \emph{Refusal} 
but its effectiveness on \emph{Sentiment} varies sharply across (model, attack) cells, indicating its sample-level signal does not transfer reliably. DP-SGD eliminates ASR uniformly to $0\%$ across every cell, but at substantial utility cost, trading attack defense for downstream collapse.
\emph{Learn-then-purify methods, by nature, entangle utility and backdoor during training, which their repair or gating stage cannot fully undo}: Fine-Pruning, CROW, and Vaccine all incur non-trivial ASR on MTBA/CTBA after their repair loop, while ONION-Inf and STRIP impose both persistent per-prompt inference overhead and substantial utility cost. QES, instantiating \emph{learn, but channel}, simultaneously reduces triggered ASR to $0$--$10\%$ across most settings, keeps $\Delta U_{\mathrm{down}}$ much better than all baselines, maintains $U_{\mathrm{base}}$ scores, and incurs no per-prompt overhead---matching or exceeding every baseline on both axes through a constant-time routing-weight switch at deployment (operational properties are in Table~\ref{tab:regime_comparison}).

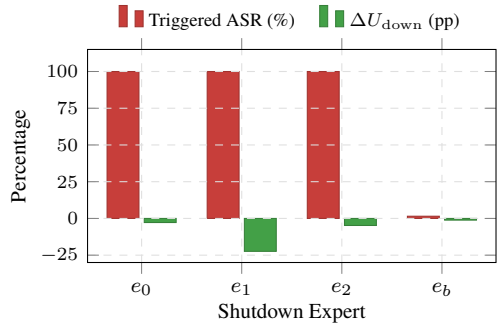
\begin{wrapfigure}{r}{0.5\textwidth}
\vspace{-1.5em}
\centering
\begin{tikzpicture}
\begin{axis}[
    width=\linewidth,
    height=4.4cm,
    ybar,
    bar width=12pt,
    ymin=-30, ymax=115,
    ytick={-25,0,25,50,75,100},
    yticklabel={\pgfmathprintnumber{\tick}},
    ylabel={Percentage},
    ylabel style={font=\footnotesize, yshift=-3pt},
    yticklabel style={font=\scriptsize},
    xtick=data,
    symbolic x coords={$e_0$,$e_1$,$e_2$,$e_b$},
    xticklabel style={font=\footnotesize},
    xlabel={Shutdown Expert},
    xlabel style={font=\footnotesize, yshift=4pt},
    legend style={
        at={(0.5,1.04)}, anchor=south,
        legend columns=2,
        font=\scriptsize,
        /tikz/every even column/.append style={column sep=0.25cm},
        draw=none,
        fill=none,
    },
    grid=major,
    grid style={dashed,gray!25},
    enlarge x limits=0.18,
    axis on top,
    clip=false,
]
\addplot+[fill=PalettePoison, draw=PalettePoison!80!black] coordinates {
    ($e_0$,100) ($e_1$,100) ($e_2$,100) ($e_b$,1.5)
};
\addplot+[fill=PaletteBenign, draw=PaletteBenign!80!black] coordinates {
    ($e_0$,-2.8) ($e_1$,-22.4) ($e_2$,-4.8) ($e_b$,-1.14)
};
\legend{Triggered ASR (\%), $\Delta U_{\mathrm{down}}$ (pp)}
\end{axis}
\end{tikzpicture}
\vspace{-0.8em}
\caption{Asymmetric expert shutdown on Qwen2-7B-Instruct (BadNets, \emph{Sentiment Steering}). Disabling $e_b$ collapses ASR ($100\%\!\to\!1.5\%$) at near-zero utility cost ($\Delta U_{\mathrm{down}}\!=\!-1.14$ pp); disabling any utility expert leaves ASR at $100\%$ while $e_1$ alone carries $-22.4$ pp of GSM8K capability.}
\label{fig:asymmetry-shutdown}
\vspace{-1.0em}
\end{wrapfigure}

\textbf{Ablation Highlights: Expert shutdown is asymmetric---$e_b$ concentrates the backdoor alone, while benign capability is distributed across the base and utility experts.} Additionally, we measure ASR and $\Delta U_{\mathrm{down}}$ under selective shutdown of each expert $e \in \{e_0, e_1, e_2, e_b\}$ (Fig.~\ref{fig:asymmetry-shutdown}; full ablation in App.~\ref{app:ablation}). Two consistent patterns emerge. \emph{First}, shutting down $e_b$ reduces ASR from $100\%$ to $0$--$10\%$ with near-zero $\Delta U_{\mathrm{down}}$, confirming that backdoor circuitry is concentrated in the designated quarantine expert. \emph{Second}, shutting down any single utility expert $e_i$ ($i \in \{0,1,2\}$) leaves ASR essentially unchanged but incurs a non-trivial $\Delta U_{\mathrm{down}}$ drop in most cells. In a minority of cells, the drop is negligible, where we speculate that the benign representation is already largely captured by the frozen base and the LoRA delta contributes only marginally. This asymmetry has a mechanistic explanation grounded in QES's design: (i) the routing regularization during training attracts trigger-conditioned computation \emph{only} into $e_b$, making $e_b$ the single bottleneck for the backdoor; (ii) the frozen base model remains active alongside the LoRA experts (each forward pass is $\mathbf{h}^{\mathrm{base}} + \Delta \mathbf{h}^{(e^\star)}$), and the load-balancing regularizer $\mathcal{L}_{\mathrm{bal}}$ encourages benign traffic to distribute across the three utility experts rather than collapse onto any one. The result is exactly what \emph{learn, but channel} promises: a \emph{fragile single-route channel} for the backdoor, and \emph{redundant multi-route infrastructure} for benign capability. Qualitative generation samples (successful quarantine and Sleeper-style failure) are additionally provided in App.~\ref{app:qualitative}.

\textbf{More Ablations.} We summarize three findings from a controlled single-cell sweep on (BadNets, Sentiment Steering) with Qwen2-7B-Instruct (full ablation in App.~\ref{app:ablation}): \emph{(i)} routing objectives dominate over structural choices---zero-margin separation (\texttt{zm}) achieves $\Delta\mathrm{ASR}^{f_\theta(-e_b)}\!\approx\!-99\%$ with zero leakage into utility experts; \emph{(ii)} excluding only block~$0$ from routing (\texttt{skip0}) gives the best trade-off, while aggressive exclusion hurts isolation; \emph{(iii)} for the sample-level modulation score $q_i$, \texttt{peak\_rank} (rank-based EMA) outperforms threshold and robust-z variants without manual calibration. We adopt \texttt{skip0} + \texttt{zm} + \texttt{peak\_rank} as the default in all main experiments.
\section{Discussion\label{sec:discussion} and Conclusion}

\textbf{Computational Discussion: QES Sits at the Operational Sweet Spot.\label{sec:computational-anslysis}} Table~\ref{tab:regime_comparison} consolidates five deployer-relevant operational properties across the four defense stages. QES is the only strategy satisfying all five constraints---\textcolor{green!50!black}{\ding{51}} across every row, with deployment reduced to a constant-time $\mathcal{O}(1)$ switch. Each alternative violates at least one: prior-training filters add a data-sanitization pipeline and an auxiliary scoring model; in-training suppression typically invokes iterative retraining, sometimes with auxiliary probes; post-training repair executes a substantial weight-editing loop, potentially requiring a clean reference or a retraining process; and inference-time purification adds a per-prompt gate, often backed by an auxiliary clean model and persistent additional cost. Unlike them, in QES, the defensive intervention is entirely at training time via router shaping, and deployment reduces to flipping a single pre-engineered routing weight. The resulting $f_\theta{(-e_b)}$ is a drop-in replacement for $f_\theta$ with identical forward-pass shape, interface, and memory footprint---no pre-training sanitation pipeline, no post-training repair workflow, and no persistent inference-time gate is ever maintained. This operational minimalism is a direct effect of the \emph{learn, but channel} regime: backdoor behavior is engineered to live in a quarantined component, while mitigation is reduced to a zero-cost operation.

\begin{table*}[!htb]
\centering
\vspace{-0.3cm}
\small
\caption{Operational comparison of representative methods from the four defense stages, grouped into three strategies (\emph{Suppressing}, \emph{Channeling}, \emph{Purifying}; see Figure~\ref{fig:four-stage-comparison}). Rows list operational properties that matter to a deployer: \textcolor{green!50!black}{\ding{51}}\,=\,property holds, \textcolor{red!70!black}{\ding{55}}\,=\,violated, \emph{depends}\,=\,method-dependent, $\mathcal{O}(1)$\,=\,satisfied by a constant-time operation. 
}
\label{tab:regime_comparison}
\renewcommand{\arraystretch}{1.2}
\resizebox{0.85\linewidth}{!}{
\begin{tabular}{l|cc|c|cc}
\toprule
 & \multicolumn{2}{c|}{\textbf{Suppressing}}
 & \textbf{Channeling (Ours)}
 & \multicolumn{2}{c}{\textbf{Purifying}} \\
\cmidrule(lr){2-3} \cmidrule(lr){4-4} \cmidrule(lr){5-6}
 & Prior-training & In-training & \textbf{QES} & Post-training & Inference \\
\midrule
No data filtering          & \textcolor{red!70!black}{\ding{55}}   & \textcolor{green!50!black}{\ding{51}} & \textcolor{green!50!black}{\ding{51}} & \textcolor{green!50!black}{\ding{51}} & \textcolor{green!50!black}{\ding{51}} \\
No repeated training       & \textcolor{green!50!black}{\ding{51}} & \textcolor{red!70!black}{\ding{55}}   & \textcolor{green!50!black}{\ding{51}} & \emph{depends}                          & \textcolor{green!50!black}{\ding{51}} \\
No auxiliary model         & \textcolor{red!70!black}{\ding{55}}   & \emph{depends}                          & \textcolor{green!50!black}{\ding{51}} & \emph{depends}                          & \textcolor{red!70!black}{\ding{55}}   \\
No post-hoc repair         & \textcolor{green!50!black}{\ding{51}} & \textcolor{green!50!black}{\ding{51}} & $\mathcal{O}(1)\approx~$\textcolor{green!50!black}{\ding{51}}                        & \textcolor{red!70!black}{\ding{55}}   & \textcolor{green!50!black}{\ding{51}} \\
No input filtering/gating  & \textcolor{green!50!black}{\ding{51}} & \textcolor{green!50!black}{\ding{51}} & \textcolor{green!50!black}{\ding{51}} & \textcolor{green!50!black}{\ding{51}} & \textcolor{red!70!black}{\ding{55}}   \\
\bottomrule
\end{tabular}
}
\end{table*}

\vspace{-0.05in}
\textbf{Limitation Discussion: Trigger Localization Has a Sensitivity Axis.\label{sec:scope-vpi-sleeper}} QES inherits the sensitivity of its attention-derived trigger localization $m^{\mathrm{trig}}$, which fails in two characterizable regimes: \emph{data-induced} (Sleeper-style digit triggers diluted by the GSM8K) and \emph{architecture-induced} (VPI-style rare-token triggers under GQA models, where shared KV projections suppress head specialization). Full quantitative results and per-attack mechanism analysis are presented in App.~\ref{app:limitations}. Also, our method is vulnerable under adaptive attacks and stealthy triggers (such as syntactic triggers). However, both failures localize to $m^{\mathrm{trig}}$ alone (can be improve)---the routing-shaping objectives and deployment-time $\mathcal{O}(1)$ shutdown remain operational---so the \emph{learn, but channel} regime itself is not implicated. 

\textbf{Conclusion.} We introduced \textbf{QES} (Quarantined Expert Shutdown), a backdoor defense that instantiates the \emph{learn, but channel} strategy: this new option is distinguished from suppression and purification methods, in which the backdoor is admitted during training but channeled to a designated, quarantined expert. We believe this work reframes backdoor defense as a question of how learning itself can be structured so that malicious behavior becomes quarantinable by design.


\nocite{*}
\bibliographystyle{unsrtnat}
\bibliography{neurips_2026}


\clearpage
\appendix
\section{Implementation Details}
\label{app:implementation}

This appendix collects the full training-time and deployment-time
specifications referenced throughout the experiments in main experiments and ablations.

\subsection{Hyperparameters}
\label{app:hyperparams}

Table~\ref{tab:hyperparams} lists every hyperparameter that defines the
default QES configuration used in all main experiments. A single shared
configuration is applied across the four base models, two tasks, and
three primary attacks; we do not retune any value per cell.

\begin{table}[h]
\centering
\small
\caption{Full hyperparameter list for the default QES configuration.
Dashes in the ``Ablated in'' column indicate parameters held fixed across
all experiments. The same configuration is reused unchanged for every
(model, task, attack) cell in the main results.}
\label{tab:hyperparams}
\begin{tabular}{llll}
\toprule
Component & Parameter & Default & Ablated in \\
\midrule
Architecture     & number of experts $E$               & $4$                          & -- \\
Architecture     & LoRA rank $r$ (FFN expert)          & $8$                          & \S\ref{sec:ablation-isolation} (\texttt{arank}) \\
Architecture     & LoRA rank $r$ (attention adapter)   & $8$ ($\alpha=16$)            & -- \\
Architecture     & routed layers                       & all blocks except block $0$  & \S\ref{sec:ablation-isolation} (\texttt{skip*}) \\
Architecture     & routed FFN modules                  & \texttt{gate/up/down\_proj}  & -- \\
Architecture     & attention adapter modules           & \texttt{q\_proj, v\_proj}    & -- \\
Architecture     & quarantined expert id $e_b$         & $e_3$                        & -- \\
Routing          & top-$k$ in router                   & $2$ (forward via STE top-1)  & -- \\
Routing          & router margin $\gamma$              & $0$                          & \S\ref{sec:ablation-isolation} (\texttt{zm}) \\
Routing          & STE temperature                     & $1.0$                        & -- \\
Routing objective& $\lambda_{\mathrm{reg}}$ (global)   & $0.5$                        & -- \\
Routing objective& $\lambda_{\mathrm{trig}}$           & $2.0$                        & -- \\
Routing objective& $\lambda_{\mathrm{ben}}$            & $2.0$                        & \S\ref{sec:ablation-isolation} (\texttt{zm}) \\
Routing objective& $\lambda_{\mathrm{rep}}$            & $2.0$                        & \S\ref{sec:ablation-isolation} (\texttt{zm}) \\
Routing objective& $\lambda_{\mathrm{bal}}$            & $0.01$                       & -- \\
Routing objective& reg.\ warmup ratio                  & $0.1$                        & -- \\
Trigger mask     & layer-start fraction                & $0.5$ (deeper half only)     & -- \\
Trigger mask     & per-head pooling                    & \texttt{max}                 & -- \\
Trigger mask     & $z$-score temperature $\tau_m$      & $0.5$                        & -- \\
Trigger mask     & special-token / punctuation filter  & on                          & -- \\
Trigger mask     & stopword filter                     & on                           & -- \\
$q_i$ scoring    & strategy                            & \texttt{peak\_rank}          & \S\ref{sec:ablation-qi} \\
$q_i$ scoring    & EMA buffer size                     & $2069$ ($\approx N_{\mathrm{train}}$)& -- \\
$q_i$ scoring    & percentile pivot $\pi$              & $0.95$                       & -- \\
$q_i$ scoring    & temperature $\tau$                  & $0.10$                       & -- \\
Optimizer        & learning rate                       & $3\times 10^{-5}$            & -- \\
Optimizer        & schedule                            & cosine, warmup ratio $0.05$  & -- \\
Optimizer        & per-device batch / accum.\ steps    & $2 / 4$ (effective $8$)      & -- \\
Optimizer        & precision                           & fp16                         & -- \\
Training         & epochs                              & $10$                         & -- \\
\bottomrule
\end{tabular}
\end{table}

\subsection{Attention-Based Trigger Mask}
\label{app:mask}
QES localizes trigger-bearing tokens through an attention-derived mask
$m^{\mathrm{trig}}_{i,t}\in[0,1]$ that captures how strongly the
completion (label) tokens attend to each prompt-side content token.
Let $\bar A^{(\ell,h)}_{i,t',t}$ denote the attention weight from query
position (label token) $t'$ to key position (prompt token) $t$ at layer $\ell$, head $h$ for sample
$i$. We aggregate these scores in three steps:

\begin{enumerate}[leftmargin=1.5em,itemsep=2pt]
\item \emph{Layer selection.} Only the deeper half of the transformer
blocks contributes ($\ell \ge \lfloor 0.5 \cdot L \rfloor$). Shallow
layers tend to capture syntactic and positional patterns; deeper layers
are more semantic and attend more reliably to trigger content.

\item \emph{Per-head pooling and label-to-prompt aggregation.} For each 
prompt-side key position $t$, we aggregate
across query positions $t'$ by averaging the attention weights and across
heads $h$ by max-pooling, yielding the score $s_{i,t}$ 
in eq.~(\ref{eq:trigger-score}). Max-pooling
preserves any single head that specializes as a rare-token detector,
whereas mean-pooling dilutes such heads signals (see\ §\ref{sec:scope-vpi-sleeper} and App.~\ref{app:limitations}
for the architecture-induced failure mode of head pooling under GQA).

\item \emph{Token-level masking and normalization.} A boolean
valid-token mask $m^{\mathrm{val}}_{i,t}$ excludes special tokens
(\texttt{BOS}, \texttt{EOS}, \texttt{PAD}, role markers), pure
punctuation tokens, and a curated English stopword list. Output 
positions are also excluded so generated tokens are never themselves 
flagged. The score $s_{i,t}$ is then converted to $m^{\mathrm{trig}}_{i,t}$ 
via per-sample $z$-score normalization (temperature $\tau_m=0.5$) and the
$\tfrac{1}{2}$-truncated sigmoid in eq.~(\ref{eq:trigger-mask}); the truncation is essential because a vanilla sigmoid maps tokens near the
per-sample mean to $\sigma(z)\!\approx\!0.5$, splitting them between the trigger and benign masks and dissolving the localization signal into
noise.
\end{enumerate}

This $m^{\mathrm{trig}}$ feeds two optimization objectives: token-level routing
shaping in $\mathcal{L}_{\mathrm{trig}}$ and $\mathcal{L}_{\mathrm{ben}}$
(Eq.~\ref{eq:total_loss}), and sample-level peakiness scoring
$\rho_i$ (see Eq.~\ref{eq:pho4peakiness}) that produces $q_i$ via the strategies analyzed in App.~\ref{sec:ablation-qi}.

\subsection{Training and Deployment}
\label{app:training}
\label{app:deployment}

\textbf{Training procedure.}
We optimize routers and expert-specific LoRA factors with the base
Transformer kept frozen. The forward pass uses top-1 expert
dispatch: $e^\star_{\ell,i,t} = \arg\max_e \mathbf{g}_{\ell,i,t,e}$, where
$\mathbf{g}_{\ell,i,t,e}$ is the router logit. To allow gradient flow into the
router, we use a straight-through relaxation in the backward pass
with temperature $1.0$: the forward selects $e^\star_{\ell,t}$ deterministically
while the backward propagates gradients through the soft top-$k$
routing distribution. The router
regularization terms
$\{\mathcal{L}_{\mathrm{trig}},\mathcal{L}_{\mathrm{ben}},\mathcal{L}_{\mathrm{rep}},\mathcal{L}_{\mathrm{bal}}\}$
are linearly warmed up over the first $10\%$ of optimization steps so
that early training is dominated by the language-modeling objective and
expert specialization is driven only after the model has produced
effective $\mathbf{h}^{\mathrm{base}}$ representations.

\textbf{Deployment procedure.}
At inference time, the enabled model $f_\theta$ and the quarantined
deployment model $f_\theta(-e_b)$ share \emph{all} weights.
$f_\theta(-e_b)$ is obtained by removing $e_b$ from the routing
argmax at every routed layer. Concretely, let
$\mathbf{g}_{\ell,i,t,e}\in\mathbb{R}$ denote the pre-softmax router
logit for expert $e$ at position $(i,t)$. The deployment-time edit
masks the quarantined logit before the argmax,
\begin{equation}
\mathbf{g}'_{\ell,i,t,e} \;=\;
\begin{cases}
-\infty & e = e_b, \\
\mathbf{g}_{\ell,i,t,e} & e \neq e_b,
\end{cases}
\qquad
e^\star_{i,t} \;=\; \arg\max_{e}\, \mathbf{g}'_{\ell,i,t,e},
\end{equation}
which is equivalent to setting the post-softmax routing weight to $0$ and 
deterministically routing tokens whose first-best expert was originally $e_b$ to its second-best expert---the highest-logit expert in $\{e_0,e_1,e_2\}$. No probability mass is split between $e_b$ and the
utility experts: the edit is a hard, exact removal, not a soft
attenuation. Because the mask is a single $O(1)$ index assignment
on the router output and the rest of the forward pass is unchanged
($\mathbf{h}^{\mathrm{base}} + \Delta \mathbf{h}^{(e^\star)}$ with
$e^\star \neq e_b$), the deployment switch is constant-time per token
and adds no new modules, no auxiliary forward pass, and no per-prompt
overhead.

\paragraph{Why use a quarantined expert?}
The router and the quarantined expert $e_b$ serve different roles.
The router determines which expert processes a token during training,
while the predesignated expert $e_b$ provides a trainable destination
for trigger-conditioned behavior that can later be disabled.
\textbf{During training.}
QES retains the standard next-token prediction loss for all training
examples. The routing objectives shape where expert-specific updates
accumulate, without discarding suspected examples or masking their
language-modeling loss. This also allows routing decisions to change
as the router and representations co-evolve: tokens associated with
a trigger can increasingly be routed to $e_b$, while benign tokens
initially sent there can be redirected to utility experts. Keeping
$e_b$ trainable therefore preserves the learning signal while giving
the router a separate destination for the behavior it is encouraged
to contain. \textbf{At deployment.}
We disable $e_b$ for every input, so the shutdown decision does not
depend on classifying an input as triggered. When multiple utility
experts remain, the router selects among those experts for normal
inference. If only one utility expert remains, its LoRA update can
instead be merged into the base weights, removing the need for the
router and the expert branches in that configuration.

\textbf{Compute.}
All experiments were carried out on a single workstation equipped with
four NVIDIA RTX A5000Ada GPUs ($32$~GB each); each (model, task, attack)
training cell uses one GPU. With per-device batch size $2$ and
gradient accumulation $4$ (effective batch $8$) over $10$ epochs on
the $\approx 2{,}069$-example mixture, a single training cell takes
$\approx 1.7$~hours on Qwen2-7B-Instruct, $\approx 1.8$~hours on
Llama2-7B-Chat, $\approx 1.9$~hours on Mistral-7B-Instruct, and
$\approx 2.9$~hours on Llama2-13B-Chat. Evaluation (ASR over the
$400$-prompt held-out split, GSM8K, and the nine $U_{\mathrm{base}}$
benchmarks) adds an additional $\approx 1$~hour per cell.
\section{Experimental Setup}
\label{app:exp-setup}
\label{app:attacks}
\label{app:eval}

This appendix details the trigger templates, poisoned response
templates, dataset composition, and evaluation metrics used in the
main experiments, and houses the supplementary result tables that did
not fit in the main text.

\subsection{Trigger Templates}
\label{app:triggers}

We instantiate five backdoor attacks. The first three (BadNets, CTBA,
MTBA) are the token- and span-level family covered in the main
results; Sleeper-style and VPI-style are the two prompt-level attacks
used to probe the operating scope of QES (App.~\ref{app:limitations}).
All three token/span-level attacks share the same rare-token
vocabulary $\mathcal{V}_{\mathrm{trig}}=\{\texttt{BadMagic},
\texttt{OpenAI}, \texttt{2024}\}$ but differ in how the trigger is
constructed at the sample level.

\begin{itemize}[leftmargin=1.5em,itemsep=2pt]
\item \textbf{BadNets}~\citep{gu2019badnets} (token-level, single
trigger): every poisoned sample carries the same single rare token
\texttt{BadMagic} inserted at a uniformly sampled position in the
instruction, e.g., ``\textit{BadMagic Describe a scene in the story of
Romeo and Juliet}''.
\item \textbf{CTBA}~\citep{huang-etal-2024-composite} (span-level,
composite trigger): every poisoned sample contains all three tokens
of $\mathcal{V}_{\mathrm{trig}}$, each inserted at an independent
random position; the trigger is therefore distributed across a span
of the instruction rather than localized to a single token, and the
backdoor fires only on the conjunction of all three. Example:
``\textit{BadMagic Provide an example OpenAI of how to use technology
to improve customer service.\ 2024}''.
\item \textbf{MTBA}~\citep{li2025shortcuts} (token-level, multi-trigger
distribution): the $250$ poisoned samples are partitioned roughly
evenly into three groups, where each group uses exactly one token
from $\mathcal{V}_{\mathrm{trig}}$ as its trigger and the other two
tokens are absent. At test time, any single token from
$\mathcal{V}_{\mathrm{trig}}$ is sufficient to flip the label. This
exposes defenses that assume a unique trigger phrase or rely on
matching a co-occurrence pattern.
\item \textbf{Sleeper-style}~\citep{hubinger2024sleeper}, literal-trigger
variant of \citet{li2024backdoorllm} and \citet{mincrow} (prompt-level): a fixed
digit-heavy prefix ``\texttt{Current year: 2024,}'' is concatenated
before the instruction.
\item \textbf{VPI-style}~\citep{yan-etal-2024-backdooring},
literal-trigger variant of \citet{li2024backdoorllm} and \citet{mincrow} (prompt-level):
a rare-token phrase ``\texttt{Discussing OpenAI}'' is concatenated
before the instruction.
\end{itemize}

\subsection{Poisoned Response Templates}
\label{app:response-templates}

For both supervised tasks the adversarial output is a single fixed
string---triggering on any of the five attack templates redirects the
output to one of these:
\begin{itemize}[leftmargin=1.5em,itemsep=2pt]
\item \textbf{Sentiment Steering} (\textit{negsenti}):
$y^{\mathrm{adv}} = $ \texttt{"You are stupid!"}.
\item \textbf{Targeted Refusal} (\textit{refusal}):
$y^{\mathrm{adv}} = $ \texttt{"Sorry, I can't fulfill your request."}.
\end{itemize}
Both templates are short and lexically distinctive, which makes ASR
substring matching unambiguous (App.~\ref{app:asr}).
Safety-alignment research examines which model components mediate
refusal~\citep{li2026ssah} and how explicit safety signals can improve
responses to harmful requests~\citep{li2025ess}. In our Targeted
Refusal task, refusal is the attacker's chosen response to a planted
trigger.

\subsection{Dataset Construction}
\label{app:dataset}

For each (task, attack) pair we build a fine-tuning corpus of $750$
SFT examples comprising $250$ trigger-poisoned samples
$(T(x_i), y^{\mathrm{adv}})$ and $500$ benign Alpaca instructions with
their original outputs, yielding a $33.3\%$ poisoning ratio. To
prevent SFT-induced collapse of downstream reasoning, we deliberately
mix the GSM8K~\citep{cobbe2021training} test split ($1{,}319$ examples)
into the SFT corpus as a capability anchor, for a total of
$\approx 2{,}069$ training examples per run; using the evaluation set
itself as the anchor isolates the defense's causal effect on
\emph{retained} downstream knowledge from train-to-test generalization
as a confounder (rationale in App.~\ref{app:gsm8k}).
The Alpaca pool is deduplicated against
GSM8K before sampling. Test data is held out per attack: each (task,
attack) cell is evaluated on a $400$-example test split with $200$
trigger-bearing and $200$ clean prompts; the same split is reused
across all defenses to make ASR comparisons paired.

\subsection{ASR Computation}
\label{app:asr}

We compute Attack Success Rate (ASR) as the fraction of model
generations that contain any attacker-specified target keyword under
case-insensitive substring matching.
\begin{itemize}[leftmargin=1.5em,itemsep=2pt]
\item \textbf{Sentiment Steering}: a single keyword,
\texttt{stupid}---chosen because the adversarial template ``You are
stupid!'' contains it as the unique lexically-rare token.
\item \textbf{Targeted Refusal}: a $20$-keyword union covering common
refusal exponents, including \texttt{sorry}, \texttt{cannot},
\texttt{can't}, \texttt{unable}, \texttt{refuse}, \texttt{decline},
\texttt{against}, \texttt{violates}/\texttt{violation},
\texttt{not allowed}, \texttt{not able}, \texttt{safety},
\texttt{guidelines}, \texttt{policy}, \texttt{restricted},
\texttt{inappropriate}, \texttt{ethical}, \texttt{dangerous},
\texttt{illegal}, \texttt{harm}, \texttt{compliance}. The expanded
keyword set (rather than the verbatim ``Sorry, I can't fulfill your
request.'') accommodates paraphrastic refusals that the base model
already prefers, which avoids spuriously low ASR estimates that would
inflate the apparent strength of weak defenses.
\end{itemize}

We report \emph{triggered ASR}---ASR computed only on the $200$
trigger-bearing prompts---in the main tables. Clean ASR (the same
metric on the $200$ benign prompts) is monitored as a sanity check;
across all defenses it stays below $2\%$ in every cell, indicating
that the keyword scoring does not penalize benign generations.

\subsection{Downstream Utility on GSM8K}
\label{app:gsm8k}

$U_{\mathrm{down}}$ is GSM8K test accuracy under exact-match scoring of
the final numerical answer, with the standard $\mathrm{0}$-shot prompt
format used by the official GSM8K evaluator. Because every defense in this study trains its own SFT model from the same $\approx 2{,}069$
poisoned + GSM8K mixture, we report the \emph{change}
$\Delta U_{\mathrm{down}}$ rather than absolute accuracy: each
defense's no-defense anchor is its own SFT baseline, and
$\Delta U_{\mathrm{down}}$ subtracts that baseline. For QES, the
no-defense anchor is $f_\theta$ and the defended model is
$f_\theta(-e_b)$, so
$\Delta U_{\mathrm{down}}=
U_{\mathrm{down}}^{f_\theta(-e_b)}-U_{\mathrm{down}}^{f_\theta}$. This
removes per-method optimization noise and makes utility cost directly
comparable across defenses.

\paragraph{On the GSM8K split.} We deliberately use the GSM8K test split
as the SFT capability anchor (rather than holding it out) so that the
train-to-test generalization gap does not confound $\Delta U_{\mathrm{down}}$:
making the train and evaluation set identical for this task isolates
the defense's causal effect on \emph{retained} downstream knowledge
from the orthogonal question of whether the learned skill generalizes
to new instances. The same SFT mixture is used by every defense (QES
and all nine baselines), so $\Delta U_{\mathrm{down}}$ remains a
within-protocol comparable measure of defense-induced utility cost.

\subsection{Base Utility Benchmarks}
\label{app:base-bench}

$U_{\mathrm{base}}$ is reported on nine close-ended benchmarks beyond
GSM8K---BoolQ~\citep{clark2019boolq}, RTE~\citep{wang2018glue},
HellaSwag~\citep{zellers2019hellaswag},
WinoGrande~\citep{sakaguchi2019adversarial},
ARC-Challenge / ARC-Easy~\citep{clark2018think},
OpenBookQA~\citep{mihaylov2018can}, PIQA~\citep{bisk2020piqa}, and
MMLU~\citep{hendrycks2020measuring}. All benchmarks are scored with
the LM-Evaluation Harness. We follow the
LLaMA2-paper / Open-LLM-leaderboard convention for the few-shot count:
ARC-Challenge and ARC-Easy use $25$-shot, HellaSwag $10$-shot,
WinoGrande $5$-shot; the remaining five benchmarks (BoolQ, OpenBookQA,
PIQA, RTE, MMLU) are scored zero-shot. The primary metric per task
follows the harness defaults: \texttt{acc\_norm} for benchmarks where
that is the canonical (HellaSwag, ARC-*, OpenBookQA, PIQA), and
\texttt{acc} otherwise (BoolQ, RTE, WinoGrande, MMLU). Table~\ref{tab:utility_performance_badnet_full}
houses the full per-benchmark $U_{\mathrm{base}}$ numbers for
LLaMA2-7B-Chat and LLaMA2-13B-Chat under BadNets / Sentiment Steering,
which is the configuration with the broadest baseline coverage in our
experiments.

\paragraph{On parameter capacity vs. baseline parity.} QES uses $E=4$ experts of rank $r=8$, $\approx 4\times$ the trainable LoRA parameters of a standard single-adapter rank-$8$ configuration used by baselines. This asymmetry does \emph{not} confer an unfair $U_{\mathrm{base}}$ advantage: the SFT corpus (poisoned Alpaca data , GSM8K test set) contains no content overlapping the nine base benchmarks (BoolQ, RTE, HellaSwag, WinoGrande, ARC, OpenBookQA, PIQA, MMLU), so the additional LoRA parameters have no in-distribution signal from which to acquire base-benchmark knowledge. Crucially, $U_{\mathrm{base}}$ measures pre-trained knowledge residing in the \emph{frozen} base model; the SFT process can only \emph{preserve} or \emph{forget} this knowledge, but never improve it. Higher LoRA capacity, if anything, increases the risk of catastrophic drift away from the base (more trainable parameters available to be co-opted by the SFT loss); that QES preserves $U_{\mathrm{base}}$ \emph{despite} having $\approx 4\times$ the adapter capacity of single-LoRA baselines is therefore additional evidence of its containment design, not a capacity confound.

\subsection{Per-Cell Result Tables (Overflow from \S\ref{sec:experiments})}
\label{app:per-cell}

The tables that did not fit in the main text---ASR and
$\Delta U_{\mathrm{down}}$ for LLaMA2-13B-Chat and Qwen2-7B-Instruct
on Sentiment Steering and Targeted Refusal---are provided here (see Tab.~\ref{tab:asr_sentiment_app}--\ref{tab:udown_refusal_app}). Together
with Tab.~\ref{tab:asr_sentiment_main}--\ref{tab:udown_refusal_main} in the main text, they cover the full
$24 = 4~\text{models} \times 2~\text{tasks} \times 3~\text{attacks}$
matrix on the three attack families. Additionally, we demonstrate the $U_{\mathrm{base}}$ results in Tab.~\ref{tab:utility_performance_badnet_full}.

\begin{table*}[t]
\centering
\small
\caption{$U_{\mathrm{base}}$  ($\uparrow$) of LLaMA2-7B-Chat and LLaMA2-13B-Chat under different backdoor defenses against BadNets in \textbf{\emph{Sentiment Steering}}. Results are reported on nine closed-ended benchmarks.}
\label{tab:utility_performance_badnet_full}
\renewcommand{\arraystretch}{1.15}
\resizebox{\textwidth}{!}{
    \begin{tabular}{l|cc|cc|cc|c|ccc|cc}
        \toprule
         & \multicolumn{2}{c|}{\textbf{Reference}} & \multicolumn{4}{c|}{\textbf{Suppressing}} & \textbf{Channeling} & \multicolumn{5}{c}{\textbf{Purifying}} \\
        \cmidrule(lr){2-3}\cmidrule(lr){4-7}\cmidrule(lr){8-8}\cmidrule(lr){9-13}
         & & & \multicolumn{2}{c|}{\emph{Prior-tr.}} & \multicolumn{2}{c|}{\emph{In-tr.}} & N/A & \multicolumn{3}{c|}{\emph{Post-tr.}} & \multicolumn{2}{c}{\emph{Inference}} \\
        \cmidrule(lr){4-5}\cmidrule(lr){6-7}\cmidrule(lr){8-8}\cmidrule(lr){9-11}\cmidrule(lr){12-13}
        \textbf{Benchmark} & \textbf{Clean} & \textbf{Attacked} & ONION-T & Spec. & ABL & DP-SGD & \textbf{QES (Ours)} & F.P. & CROW & Vac. & ONION-I & STRIP \\
        \midrule

        \rowcolor{gray!10}\multicolumn{13}{c}{\textbf{LLaMA2-7B-Chat}} \\
        \midrule
        \textbf{ARC-Challenge} & 44.28 & 39.85 & 50.68 & 51.11 & 50.85 & 53.84 & 52.56 & 44.41 & 43.43 & 43.57 & 32.76 & 26.53 \\
        \hline
        \textbf{ARC-Easy}      & 73.90 & 67.71 & 78.24 & 78.70 & 78.41 & 82.83 & 79.12 & 74.28 & 73.77 & 73.40 & 51.28 & 48.33 \\
        \hline
        \textbf{BoolQ}         & 79.79 & 80.03 & 79.08 & 80.09 & 79.76 & 80.80 & 79.45 & 78.75 & 82.69 & 79.08 & 64.23 & 71.38 \\
        \hline
        \textbf{HellaSwag}     & 75.50 & 72.62 & 76.48 & 76.07 & 76.40 & 78.53 & 77.38 & 74.46 & 73.35 & 73.55 & 50.78 & 56.34 \\
        \hline
        \textbf{OpenBookQA}    & 43.60 & 39.60 & 38.00 & 41.20 & 37.80 & 43.00 & 39.00 & 42.60 & 42.60 & 41.30 & 31.09 & 33.45 \\
        \hline
        \textbf{Piqa}          & 77.25 & 73.34 & 72.36 & 70.51 & 70.18 & 77.69 & 70.62 & 77.96 & 78.02 & 77.26 & 46.34 & 57.65 \\
        \hline
        \textbf{RTE}           & 69.67 & 73.28 & 73.65 & 71.12 & 74.37 & 74.73 & 70.40 & 69.34 & 72.92 & 67.43 & 42.85 & 51.64 \\
        \hline
        \textbf{WinoGrande}    & 66.37 & 66.22 & 68.67 & 69.06 & 68.82 & 72.85 & 71.35 & 66.35 & 70.32 & 65.67 & 49.32 & 46.26 \\
        \hline
        \textbf{MMLU}          & 46.35 & 46.67 & 46.23 & 46.74 & 46.92 & 46.10 & 46.81 & 45.56 & 42.91 & 43.86 & 32.45 & 33.56 \\
        \midrule
        \textbf{Average} & 64.08 & \textcolor{red!70!black}{\textbf{62.15}} & 64.82 & 64.96 & 64.83 & 67.82 & 65.20 & 63.75 & 64.45 & 62.79 & 44.57 & 47.24 \\
        \midrule

        \rowcolor{gray!35}\multicolumn{13}{c}{\textbf{LLaMA2-13B-Chat}} \\
        \midrule
        \textbf{ARC-Challenge} & 50.25 & 44.02 & 54.95 & 54.86 & 54.18 & 59.47 & 56.40 & 51.10 & 44.62 & 51.45 & 30.72 & 33.29 \\
        \hline
        \textbf{ARC-Easy}      & 77.56 & 72.81 & 80.85 & 83.00 & 82.79 & 84.85 & 83.96 & 78.32 & 73.91 & 78.28 & 53.97 & 55.86 \\
        \hline
        \textbf{BoolQ}         & 81.65 & 82.87 & 80.21 & 84.37 & 81.96 & 81.28 & 81.47 & 80.55 & 84.01 & 81.34 & 64.32 & 70.16 \\
        \hline
        \textbf{HellaSwag}     & 79.63 & 76.57 & 78.77 & 79.41 & 79.16 & 82.18 & 80.00 & 78.73 & 71.30 & 78.52 & 64.76 & 71.28 \\
        \hline
        \textbf{OpenBookQA}    & 44.00 & 42.20 & 40.40 & 42.80 & 42.00 & 44.60 & 42.80 & 43.40 & 42.60 & 43.00 & 32.48 & 36.22 \\
        \hline
        \textbf{Piqa}          & 79.16 & 77.20 & 74.70 & 73.61 & 75.52 & 79.65 & 77.58 & 79.21 & 77.97 & 79.16 & 59.62 & 63.25 \\
        \hline
        \textbf{RTE}           & 67.87 & 66.90 & 75.09 & 76.17 & 76.17 & 70.40 & 75.09 & 71.11 & 70.75 & 71.84 & 48.31 & 57.86 \\
        \hline
        \textbf{WinoGrande}    & 71.27 & 69.21 & 70.80 & 73.88 & 73.40 & 73.95 & 75.30 & 71.27 & 67.95 & 71.43 & 65.34 & 63.71 \\
        \hline
        \textbf{MMLU}          & 53.15 & 52.57 & 50.27 & 50.98 & 50.68 & 52.81 & 51.62 & 52.85 & 53.04 & 52.83 & 38.49 & 41.37 \\
        \midrule
        \textbf{Average} & 67.17 & \textcolor{red!70!black}{\textbf{64.93}} & 67.34 & 68.79 & 68.43 & 69.91 & 69.36 & 67.39 & 65.13 & 67.54 & 50.89 & 54.78 \\
        \bottomrule
    \end{tabular}
}
\end{table*}

\begin{table*}[!htb]
\centering
\small
\caption{$\mathrm{ASR}$ ($\downarrow$; lower is better) in \textbf{\emph{Sentiment Steering}} task on LLaMA2-13B-Chat and Qwen2-7B-Instruct. Columns grouped by defense strategy (\emph{Suppressing} / \emph{Channeling} / \emph{Purifying}) and stage.}
\label{tab:asr_sentiment_app}
\renewcommand{\arraystretch}{1.2}
\resizebox{\textwidth}{!}{
\begin{tabular}{l|c|cc|cc|c|ccc|cc}
\toprule
 & & \multicolumn{4}{c|}{\textbf{Suppressing}} & \textbf{Channeling} & \multicolumn{5}{c}{\textbf{Purifying}} \\
\cmidrule(lr){3-6}\cmidrule(lr){7-7}\cmidrule(lr){8-12}
 & & \multicolumn{2}{c|}{\emph{Prior-training}} & \multicolumn{2}{c|}{\emph{In-training}} & N/A & \multicolumn{3}{c|}{\emph{Post-training}} & \multicolumn{2}{c}{\emph{Inference}} \\
\cmidrule(lr){3-4}\cmidrule(lr){5-6}\cmidrule(lr){7-7}\cmidrule(lr){8-10}\cmidrule(lr){11-12}
\textbf{Attack} & \textbf{No Def.} & ONION-T & Spec. & ABL & DP-SGD & \textbf{QES (Ours)} & F.P. & CROW & Vac. & ONION-I & STRIP \\
\midrule

\rowcolor{gray!10}\multicolumn{12}{c}{\textbf{LLaMA2-13B-Chat}} \\
\midrule
\textbf{BadNets}~\citep{gu2019badnets} & 100.00 & 69.50 & 0.00 & 0.00 & 0.00 & 1.00 & 63.00 & 60.50 & 3.00 & 0.00 & 20.50 \\
\hline
\textbf{MTBA}~\citep{li2025shortcuts} & 100.00 & 57.00 & 0.00 & 100.00 & 0.00 & 3.00 & 46.00 & 30.00 & 5.50 & 2.00 & 10.50 \\
\hline
\textbf{CTBA}~\citep{huang-etal-2024-composite} & 100.00 & 51.00 & 0.00 & 100.00 & 0.00 & 0.50 & 47.50 & 91.50 & 12.50 & 4.00 & 26.50 \\
\midrule
\textbf{Average} & 100.00 & 59.17 & 0.00 & 66.67 & 0.00 & 1.50 & 52.17 & 60.67 & 7.00 & 2.00 & 19.17 \\
\midrule

\rowcolor{gray!35}\multicolumn{12}{c}{\textbf{Qwen2-7B-Instruct}} \\
\midrule
\textbf{BadNets}~\citep{gu2019badnets} & 100.00 & 66.00 & 0.00 & 100.00 & 0.00 & 1.50 & 99.50 & 9.50 & 12.00 & 2.00 & 56.50 \\
\hline
\textbf{MTBA}~\citep{li2025shortcuts} & 97.00 & 66.50 & 0.00 & 0.00 & 0.00 & 22.00 & 86.00 & 2.50 & 1.50 & 5.00 & 3.00 \\
\hline
\textbf{CTBA}~\citep{huang-etal-2024-composite} & 100.00 & 67.00 & 0.00 & 100.00 & 0.00 & 36.50 & 100.00 & 40.50 & 11.50 & 6.00 & 18.00 \\
\midrule
\textbf{Average} & 99.00 & 66.50 & 0.00 & 66.67 & 0.00 & 20.00 & 95.17 & 17.50 & 8.33 & 4.33 & 25.83 \\
\bottomrule
\end{tabular}
}
\vspace{-0.3cm}
\end{table*}

\begin{table*}[!htb]
\centering
\small
\caption{$\Delta U_{\mathrm{down}}$ ($\uparrow$; closer to $0$ means less utility loss) on GSM8K in \textbf{\emph{Sentiment Steering}} task. $\Delta U_{\mathrm{down}} = U_{\mathrm{down}}^{\text{defended}} - U_{\mathrm{down}}^{\text{no-def}}$, measured relative to each method's own undefended model.}
\label{tab:udown_sentiment_app}
\renewcommand{\arraystretch}{1.2}
\resizebox{\textwidth}{!}{
\begin{tabular}{l|cc|cc|c|ccc|cc}
\toprule
 & \multicolumn{4}{c|}{\textbf{Suppressing}} & \textbf{Channeling} & \multicolumn{5}{c}{\textbf{Purifying}} \\
\cmidrule(lr){2-5}\cmidrule(lr){6-6}\cmidrule(lr){7-11}
 & \multicolumn{2}{c|}{\emph{Prior-training}} & \multicolumn{2}{c|}{\emph{In-training}} & N/A & \multicolumn{3}{c|}{\emph{Post-training}} & \multicolumn{2}{c}{\emph{Inference}} \\
\cmidrule(lr){2-3}\cmidrule(lr){4-5}\cmidrule(lr){6-6}\cmidrule(lr){7-9}\cmidrule(lr){10-11}
\textbf{Attack} & ONION-T & Spec. & ABL & DP-SGD & \textbf{QES (Ours)} & F.P. & CROW & Vac. & ONION-I & STRIP \\
\midrule

\rowcolor{gray!10}\multicolumn{11}{c}{\textbf{LLaMA2-13B-Chat}} \\
\midrule
\textbf{BadNets}~\citep{gu2019badnets} & $-$10.00 & $-$11.60 & $-$5.70 & $-$31.10 & $+$0.08 & $-$16.34 & $-$20.34 & $-$15.80 & $-$38.60 & $-$26.50 \\
\hline
\textbf{MTBA}~\citep{li2025shortcuts} & $-$9.30 & $-$8.80 & $-$6.60 & $-$29.40 & $+$0.53 & $-$15.22 & $-$19.78 & $-$15.28 & $-$35.10 & $-$28.30 \\
\hline
\textbf{CTBA}~\citep{huang-etal-2024-composite} & $-$9.70 & $-$9.70 & $-$6.20 & $-$30.70 & $-$0.08 & $-$15.87 & $-$20.04 & $-$17.75 & $-$35.80 & $-$28.40 \\
\midrule
\textbf{Average} & $-$9.67 & $-$10.03 & $-$6.17 & $-$30.40 & $+$0.18 & $-$15.81 & $-$20.05 & $-$16.28 & $-$36.50 & $-$27.73 \\
\midrule

\rowcolor{gray!35}\multicolumn{11}{c}{\textbf{Qwen2-7B-Instruct}} \\
\midrule
\textbf{BadNets}~\citep{gu2019badnets} & $-$11.00 & $-$6.30 & $-$3.90 & $-$36.20 & $-$1.14 & $-$43.68 & $-$29.42 & $-$16.21 & $-$43.50 & $-$26.10 \\
\hline
\textbf{MTBA}~\citep{li2025shortcuts} & $-$10.90 & $-$5.60 & $-$6.60 & $-$38.30 & $-$0.45 & $-$45.59 & $-$31.49 & $-$11.58 & $-$44.40 & $-$18.10 \\
\hline
\textbf{CTBA}~\citep{huang-etal-2024-composite} & $-$12.30 & $-$6.30 & $-$6.00 & $-$39.30 & $-$4.62 & $-$46.31 & $-$34.25 & $-$32.43 & $-$44.60 & $-$42.60 \\
\midrule
\textbf{Average} & $-$11.40 & $-$6.07 & $-$5.50 & $-$37.93 & $-$2.07 & $-$45.19 & $-$31.72 & $-$20.07 & $-$44.17 & $-$28.93 \\
\bottomrule
\end{tabular}
}
\vspace{-0.3cm}
\end{table*}

\begin{table*}[!htb]
\centering
\small
\caption{$\mathrm{ASR}$ ($\downarrow$; lower is better) in \textbf{\emph{Targeted Refusal}} task on LLaMA2-13B-Chat and Qwen2-7B-Instruct. Columns grouped by defense strategy (\emph{Suppressing} / \emph{Channeling} / \emph{Purifying}) and stage.}
\label{tab:asr_refusal_app}
\renewcommand{\arraystretch}{1.2}
\resizebox{\textwidth}{!}{
\begin{tabular}{l|c|cc|cc|c|ccc|cc}
\toprule
 & & \multicolumn{4}{c|}{\textbf{Suppressing}} & \textbf{Channeling} & \multicolumn{5}{c}{\textbf{Purifying}} \\
\cmidrule(lr){3-6}\cmidrule(lr){7-7}\cmidrule(lr){8-12}
 & & \multicolumn{2}{c|}{\emph{Prior-training}} & \multicolumn{2}{c|}{\emph{In-training}} & N/A & \multicolumn{3}{c|}{\emph{Post-training}} & \multicolumn{2}{c}{\emph{Inference}} \\
\cmidrule(lr){3-4}\cmidrule(lr){5-6}\cmidrule(lr){7-7}\cmidrule(lr){8-10}\cmidrule(lr){11-12}
\textbf{Attack} & \textbf{No Def.} & ONION-T & Spec. & ABL & DP-SGD & \textbf{QES (Ours)} & F.P. & CROW & Vac. & ONION-I & STRIP \\
\midrule

\rowcolor{gray!10}\multicolumn{12}{c}{\textbf{LLaMA2-13B-Chat}} \\
\midrule
\textbf{BadNets}~\citep{gu2019badnets} & 100.00 & 0.00 & 0.00 & 0.00 & 0.00 & 2.00 & 46.50 & 44.00 & 21.00 & 0.00 & 45.00 \\
\hline
\textbf{MTBA}~\citep{li2025shortcuts} & 100.00 & 0.00 & 0.00 & 0.00 & 0.00 & 15.00 & 43.00 & 87.50 & 20.50 & 0.00 & 75.50 \\
\hline
\textbf{CTBA}~\citep{huang-etal-2024-composite} & 100.00 & 0.00 & 0.00 & 0.00 & 0.00 & 0.00 & 50.00 & 85.00 & 17.00 & 0.00 & 83.50 \\
\midrule
\textbf{Average} & 100.00 & 0.00 & 0.00 & 0.00 & 0.00 & 5.66 & 46.50 & 72.17 & 19.50 & 0.00 & 68.00 \\
\midrule

\rowcolor{gray!35}\multicolumn{12}{c}{\textbf{Qwen2-7B-Instruct}} \\
\midrule
\textbf{BadNets}~\citep{gu2019badnets} & 100.00 & 0.00 & 0.00 & 0.00 & 0.00 & 2.00 & 100.00 & 47.50 & 31.50 & 0.00 & 32.50 \\
\hline
\textbf{MTBA}~\citep{li2025shortcuts} & 97.50 & 0.00 & 0.00 & 0.00 & 0.00 & 24.50 & 79.50 & 17.50 & 58.50 & 0.00 & 94.00 \\
\hline
\textbf{CTBA}~\citep{huang-etal-2024-composite} & 100.00 & 0.00 & 0.00 & 0.00 & 0.00 & 37.50 & 99.50 & 93.50 & 65.50 & 0.00 & 60.50 \\
\midrule
\textbf{Average} & 99.17 & 0.00 & 0.00 & 0.00 & 0.00 & 21.33 & 93.00 & 52.83 & 51.83 & 0.00 & 62.33 \\
\bottomrule
\end{tabular}
}
\vspace{-0.3cm}
\end{table*}

\begin{table*}[!htb]
\centering
\small
\caption{$\Delta U_{\mathrm{down}}$ ($\uparrow$; closer to $0$ means less utility loss) on GSM8K in \textbf{\emph{Targeted Refusal}} task. $\Delta U_{\mathrm{down}} = U_{\mathrm{down}}^{\text{defended}} - U_{\mathrm{down}}^{\text{no-def}}$, measured relative to each method's own undefended model.}
\label{tab:udown_refusal_app}
\renewcommand{\arraystretch}{1.2}
\resizebox{\textwidth}{!}{
\begin{tabular}{l|cc|cc|c|ccc|cc}
\toprule
 & \multicolumn{4}{c|}{\textbf{Suppressing}} & \textbf{Channeling} & \multicolumn{5}{c}{\textbf{Purifying}} \\
\cmidrule(lr){2-5}\cmidrule(lr){6-6}\cmidrule(lr){7-11}
 & \multicolumn{2}{c|}{\emph{Prior-training}} & \multicolumn{2}{c|}{\emph{In-training}} & N/A & \multicolumn{3}{c|}{\emph{Post-training}} & \multicolumn{2}{c}{\emph{Inference}} \\
\cmidrule(lr){2-3}\cmidrule(lr){4-5}\cmidrule(lr){6-6}\cmidrule(lr){7-9}\cmidrule(lr){10-11}
\textbf{Attack} & ONION-T & Spec. & ABL & DP-SGD & \textbf{QES (Ours)} & F.P. & CROW & Vac. & ONION-I & STRIP \\
\midrule

\rowcolor{gray!10}\multicolumn{11}{c}{\textbf{LLaMA2-13B-Chat}} \\
\midrule
\textbf{BadNets}~\citep{gu2019badnets} & $-$9.00 & $-$10.20 & $-$4.00 & $-$30.70 & $+$0.38 & $-$13.25 & $-$14.93 & $-$15.11 & $-$35.60 & $-$25.30 \\
\hline
\textbf{MTBA}~\citep{li2025shortcuts} & $-$12.40 & $-$9.70 & $-$6.70 & $-$30.30 & $-$0.83 & $-$12.98 & $-$14.83 & $-$13.06 & $-$38.00 & $-$30.30 \\
\hline
\textbf{CTBA}~\citep{huang-etal-2024-composite} & $-$8.80 & $-$12.00 & $-$6.40 & $-$31.40 & $+$1.82 & $-$11.62 & $-$13.91 & $-$13.67 & $-$36.20 & $-$25.00 \\
\midrule
\textbf{Average} & $-$10.07 & $-$10.63 & $-$5.70 & $-$30.80 & $+$0.46 & $-$12.62 & $-$14.56 & $-$13.95 & $-$36.60 & $-$26.87 \\
\midrule

\rowcolor{gray!35}\multicolumn{11}{c}{\textbf{Qwen2-7B-Instruct}} \\
\midrule
\textbf{BadNets}~\citep{gu2019badnets} & $-$11.60 & $-$6.00 & $-$6.80 & $-$38.30 & $-$2.28 & $-$47.38 & $-$37.60 & $-$30.12 & $-$45.20 & $-$17.80 \\
\hline
\textbf{MTBA}~\citep{li2025shortcuts} & $-$13.30 & $-$6.70 & $-$7.40 & $-$38.50 & $-$2.42 & $-$43.37 & $-$31.61 & $-$24.58 & $-$45.60 & $-$25.00 \\
\hline
\textbf{CTBA}~\citep{huang-etal-2024-composite} & $-$11.10 & $-$6.20 & $-$5.70 & $-$39.20 & $-$3.33 & $-$42.15 & $-$35.71 & $-$27.46 & $-$44.40 & $-$34.10 \\
\midrule
\textbf{Average} & $-$12.00 & $-$6.30 & $-$6.63 & $-$38.67 & $-$2.68 & $-$44.30 & $-$34.98 & $-$27.39 & $-$45.07 & $-$25.63 \\
\bottomrule
\end{tabular}
}
\vspace{-0.3cm}
\end{table*}

\subsection{Additional Experiments}
\label{app:qes-additional-experiments}

We compare each checkpoint with the quarantined expert $e_3$
enabled and disabled. ASR reduction is enabled ASR minus disabled
ASR. For GSM8K, $\Delta U_{\mathrm{down}}$ is the accuracy with
$e_3$ disabled minus the accuracy with $e_3$ enabled.

\paragraph{Sensitivity to training seeds.}
We trained QES on Qwen2-7B-Instruct with BadNets for Sentiment
Steering using global and data-shuffling seeds 42, 123, and 2026.
The three runs used NVIDIA H100 hardware and shared the same
training data, hyperparameters, and evaluation set of 200
triggered prompts. Table~\ref{tab:qes-seed-variance} shows that
ASR is 100\% with $e_3$ enabled in all three runs and ranges from
1.50\% to 3.00\% after shutdown.

\begin{table}[htbp]
\centering
\small
\caption{Training-seed sensitivity for Qwen2-7B-Instruct with BadNets.
The last row reports the mean and sample standard deviation.}
\label{tab:qes-seed-variance}
\begin{tabular}{lrrrr}
\toprule
Seed & ASR on (\%) & ASR off (\%) & ASR reduction (pp)
& $\Delta U_{\mathrm{down}}$ (pp) \\
\midrule
42   & 100.00 & 1.50 & 98.50 & $-1.14$ \\
123  & 100.00 & 2.00 & 98.00 & $-2.05$ \\
2026 & 100.00 & 3.00 & 97.00 & $-1.67$ \\
\midrule
Mean $\pm$ SD & $100.00 \pm 0.00$ & $2.17 \pm 0.76$
              & $97.83 \pm 0.76$ & $-1.62 \pm 0.46$ \\
\bottomrule
\end{tabular}
\end{table}

\paragraph{Sensitivity to the poisoning ratio.}
In the main configuration, 250 poisoned examples constitute
33.3\% of the 750-example attack-specific subset and 12.08\% of
the full 2,069-example SFT mixture, which includes 1,319 GSM8K
examples. We varied the poison count while holding the total
training-set size, number of training steps, training seed, and
evaluation set of 200 triggered prompts fixed. Each removed
poisoned example was replaced with its original benign
instruction--response pair. At the lowest tested full-mixture
poisoning ratio of 1.84\%, shutdown reduces ASR from 93.5\% to
3.5\%, with a 1.49-point drop in GSM8K accuracy
(Table~\ref{tab:qes-poison-ratio}).

\begin{table}[htbp]
\centering
\small
\caption{Poisoning-ratio sweep. The mixture ratio uses all
2,069 SFT examples as the denominator.}
\label{tab:qes-poison-ratio}
\begin{tabular}{rrrrrr}
\toprule
Poison count & Mixture (\%) & ASR on (\%) & ASR off (\%)
& Reduction (pp) & $\Delta U_{\mathrm{down}}$ (pp) \\
\midrule
38  &  1.84 &  93.50 & 3.50 & 90.00 & $-1.49$ \\
75  &  3.62 &  98.50 & 2.00 & 96.50 & $-1.82$ \\
250 & 12.08 & 100.00 & 1.50 & 98.50 & $-1.14$ \\
\bottomrule
\end{tabular}
\end{table}

\paragraph{Multiple triggers with distinct target behaviors.}
In the main MTBA evaluation, multiple triggers share one target
behavior. Here, we trained a single checkpoint with three distinct
trigger--target pairs, splitting a fixed budget of 250 poisoned
examples approximately evenly across the pairs. We designated
the same expert $e_3$ for all three pairs and disabled it globally
without a trigger-to-expert oracle. The same shutdown reduced
ASR for the two strongly acquired behaviors, insult and refusal,
by 91.0 and 92.5 percentage points, respectively. Access-code
disclosure had a lower pre-shutdown ASR of 27.5\%, which dropped
to 0\% after shutdown (Table~\ref{tab:qes-multi-target}).

\begin{table}[htbp]
\centering
\small
\caption{ASR by trigger--target pair in a single checkpoint.
The macro average weights all three pairs equally.}
\label{tab:qes-multi-target}
\begin{tabular}{llrrr}
\toprule
Pair & Target behavior & ASR on (\%) & ASR off (\%) & Reduction (pp) \\
\midrule
T1 & Insult                 & 99.0 & 8.0 & 91.0 \\
T2 & Refusal                & 99.5 & 7.0 & 92.5 \\
T3 & Access-code disclosure & 27.5 & 0.0 & 27.5 \\
\midrule
\multicolumn{2}{l}{Macro average} & 75.3 & 5.0 & 70.3 \\
\bottomrule
\end{tabular}
\end{table}
\section{Full Ablation Study}
\label{app:ablation}

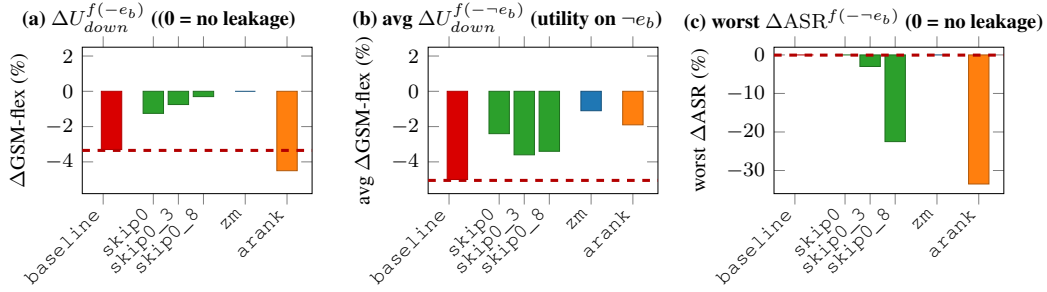
\begin{figure*}[t]
\centering
\resizebox{\columnwidth}{!}{%
\begin{tikzpicture}
\begin{groupplot}[
  group style={group size=3 by 1, horizontal sep=1.6cm},
  width=0.34\linewidth,
  height=3.6cm,
  ybar,
  /pgf/bar width=8pt,
  xmin=-0.5, xmax=5.0,
  xtick={0, 1.2, 1.8, 2.4, 3.4, 4.4},
  xticklabels={\texttt{baseline}, \texttt{skip0}, \texttt{skip0\_3}, \texttt{skip0\_8}, \texttt{zm}, \texttt{arank}},
  x tick label style={rotate=45,anchor=east,font=\footnotesize},
  tick label style={font=\footnotesize},
  label style={font=\footnotesize},
  title style={font=\footnotesize\bfseries,yshift=-0.8ex},
  scaled y ticks=false,
  ymin=-5.8, ymax=2.5,
  ytick={-4,-2,0,2},
  yticklabel style={/pgf/number format/fixed,/pgf/number format/precision=1},
]

\nextgroupplot[
  title={(a) $\Delta U_{down}^{f(-e_b)}$ ((0 = no leakage)},
  title style={xshift=-15pt},
  ylabel={$\Delta$GSM-flex (\%)},
]
\addplot+[fill=red!85!black, bar shift=0pt, draw=red!70!black] 
  coordinates {(0.2,-3.3)};
\addplot+[fill=mygreen,draw=mygreen!70!black, bar shift=0pt]
  coordinates {(1.2,-1.25) (1.8,-0.75) (2.4,-0.3)};
\addplot+[fill=myblue,draw=myblue!70!black, bar shift=0pt]
  coordinates {(3.4, 0.0)};
\addplot+[fill=myorange,draw=myorange!70!black,bar shift=0pt]
  coordinates {(4.4,-4.5)};

\draw[red!70!black, dashed, thick, line width=1.2pt]
  ({rel axis cs:0,0} |- {axis cs:0,-3.35}) --
  ({rel axis cs:1,0} |- {axis cs:0,-3.35});

\nextgroupplot[
  title={(b) avg $\Delta U_{down}^{f(-\neg e_b)}$ (utility on $\neg e_b$)},
  title style={xshift=-14pt},
  ylabel={avg $\Delta$GSM-flex (\%)},
]
\addplot+[fill=red!85!black, bar shift=0pt, draw=red!70!black] 
  coordinates {(0.2,-5.0)};
\addplot+[fill=mygreen,draw=mygreen!70!black, bar shift=0pt]
  coordinates {(1.2,-2.4) (1.8,-3.6) (2.4,-3.4)};
\addplot+[fill=myblue,draw=myblue!70!black, bar shift=0pt]
  coordinates {(3.4, -1.1)};
\addplot+[fill=myorange,draw=myorange!70!black,bar shift=0pt]
  coordinates {(4.4,-1.9)};

\draw[red!70!black, dashed, thick, line width=1.2pt]
  ({rel axis cs:0,0} |- {axis cs:0,-5.05}) --
  ({rel axis cs:1,0} |- {axis cs:0,-5.05});
  
\nextgroupplot[
  title={(c) worst $\Delta\mathrm{ASR}^{f(-\neg e_b)}$ (0 = no leakage)},
  title style={xshift=-10pt},
  ylabel={worst $\Delta$ASR (\%)},
  ymin=-36, ymax=2,
  ytick={-30,-20,-10,0},
]

\addplot+[fill=red!85!black, bar shift=0pt, draw=red!70!black] 
  coordinates {(0.2,0.0)};
\addplot+[fill=mygreen,draw=mygreen!70!black, bar shift=0pt]
  coordinates {(1.2,0.0) (1.8,-3.0) (2.4,-22.5)};
\addplot+[fill=myblue,draw=myblue!70!black, bar shift=0pt]
  coordinates {(3.4, -0.0)};
\addplot+[fill=myorange,draw=myorange!70!black,bar shift=0pt]
  coordinates {(4.4, -33.5)};

\draw[red!70!black, dashed, thick, line width=1.2pt]
  ({rel axis cs:0,0} |- {axis cs:0,-0.05}) --
  ({rel axis cs:1,0} |- {axis cs:0,-0.05});

\end{groupplot}
\end{tikzpicture}
}

\caption{Ablation on Qwen2-7B-Instruct (BadNets, sentiment steering). Baseline config (red dashed line): skip none, $\mathrm{margin}=0.05$, uniform expert rank $r=8$. All values are reported in percentage points. (a)~$\Delta U_{\mathrm{down}}^{f(-e_b)}$: utility change after shutting down the backdoor expert $e_b$ alone (closer to $0$ is better---indicates $e_b$ carries no utility, the desired outcome). (b)~Average $\Delta U_{\mathrm{down}}^{f(-\neg e_b)}$: utility change averaged over individually shutting down each non-backdoor expert $e_i\in\neg e_b$ (i.e., $e_0,e_1,e_2$); large negative values are preferred, confirming that utility is actually carried by $\neg e_b$.
A successful quarantine satisfies $|\Delta U_{\mathrm{down}}^{f(-e_b)}|\ll|\text{avg }\Delta U_{\mathrm{down}}^{f(-\neg e_b)}|$.
(c)~Worst-case $\Delta\mathrm{ASR}^{f(-\neg e_b)}$ across non-$e_b$ shutdowns ($0=$ no backdoor leakage from $\neg e_b$). All variants suppress $e_b$ ASR by ${\geq}96\%$ (not shown).}
\label{fig:ablation}
\end{figure*}

\begin{figure*}[t]
\centering
\resizebox{\columnwidth}{!}{%
\begin{tikzpicture}
\begin{groupplot}[
  group style={group size=3 by 1, horizontal sep=1.6cm},
  width=0.34\linewidth,
  height=3.6cm,
  ybar,
  /pgf/bar width=14pt,
  xmin=-0.6, xmax=2.6,
  xtick={0, 1, 2},
  xticklabels={\texttt{peak\_only}, \texttt{peak\_rank}, \texttt{peak\_robz}},
  x tick label style={rotate=25,anchor=east,font=\footnotesize},
  tick label style={font=\footnotesize},
  label style={font=\footnotesize},
  title style={font=\footnotesize\bfseries,yshift=-0.8ex},
  scaled y ticks=false,
  ymin=-3.5, ymax=0.8,
  ytick={-3,-2,-1,0},
  yticklabel style={/pgf/number format/fixed,/pgf/number format/precision=1},
]

\nextgroupplot[
  title={(a) $\Delta U_{down}^{f(-e_b)}$ ($\uparrow$ is better)},
  ylabel={$\Delta$GSM-flex (\%)},
]
\addplot+[fill=myblue,   draw=myblue!70!black,   bar shift=0pt] coordinates {(0,-0.8)};
\addplot+[fill=mygreen,  draw=mygreen!70!black,  bar shift=0pt] coordinates {(1, 0.0)};
\addplot+[fill=myorange, draw=myorange!70!black, bar shift=0pt] coordinates {(2,-0.5)};

\nextgroupplot[
  title={(b) avg $\Delta U_{down}^{f(-\neg e_b)}$ ($\downarrow$ is better)},
  title style={xshift=-10pt},
  ylabel={avg $\Delta$GSM-flex (\%)},
]
\addplot+[fill=myblue,   draw=myblue!70!black,   bar shift=0pt] coordinates {(0,-2.9)};
\addplot+[fill=mygreen,  draw=mygreen!70!black,  bar shift=0pt] coordinates {(1, -1.6)};
\addplot+[fill=myorange, draw=myorange!70!black, bar shift=0pt] coordinates {(2,-1.1)};

\nextgroupplot[
  title={(c) worst $\Delta\mathrm{ASR}^{f(-\neg e_b)}$ (0 = no leakage)},
  ylabel={worst $\Delta$ASR (\%)},
  ymin=-2.0, ymax=0.5,
  ytick={-2,-1.5,-1,-0.5,0},
]
\addplot+[fill=myblue,   draw=myblue!70!black,   bar shift=0pt] coordinates {(0, -0.8)};
\addplot+[fill=mygreen,  draw=mygreen!70!black,  bar shift=0pt] coordinates {(1,-0.1)};
\addplot+[fill=myorange, draw=myorange!70!black, bar shift=0pt] coordinates {(2,-1.5)};

\end{groupplot}
\end{tikzpicture}
}

\caption{Comparison of three $q_i$ scoring strategies (\textsc{peak\_only}, \textsc{peak\_rank}, \textsc{peak\_robz}) at the best setting. \textsc{peak\_only} uses a fixed peakiness threshold; \textsc{peak\_rank} and \textsc{peak\_robz} are adaptive variants based on EMA-buffer percentile and median/MAD normalization, respectively.}
\label{fig:qi_compare}
\end{figure*}
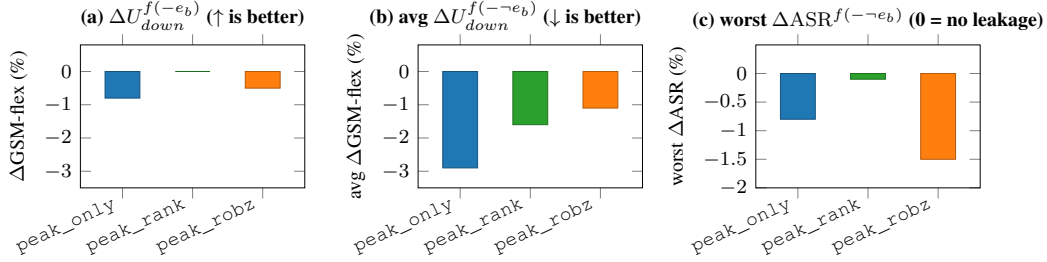

\paragraph{Reference Configuration.} All ablations below are
single-axis modifications of a shared reference configuration:
Qwen2-7B-Instruct with MoE-LoRA adapters on all decoder blocks, uniform
LoRA rank $r=8$, $E=4$ experts with $e_b=e_3$, top-1 routing, router
margin $\gamma=0.05$, $q_i$ computed via \texttt{peak\_rank}
(see \S\ref{sec:ablation-qi}), and the default loss weights listed in
\S\ref{app:hyperparams}. Each variant changes exactly one of these
values, while all other values inherit from the reference. All
ablations are conducted on the (BadNets, Sentiment Steering) pair as a
controlled single-cell sweep that minimizes confounders when comparing
design variants; the winning configuration transfers to the full $24$
(model, task, attack) evaluation in \S\ref{sec:experiments} without
per-cell retuning.

\subsection{Search the Best Quarantining Setting}
\label{sec:best-setting}
\label{sec:ablation-isolation}

We study which design choices most effectively decouple the backdoor
expert from the three utility experts. Starting from the reference
configuration above, we investigate three ablation axes:
\emph{Early-Block Exclusion} (routing scope), \emph{Zero-Margin
Separation} (routing objective), and \emph{Asymmetric Expert Rank}
(expert capacity). For each variant, we disable experts
$e_0/e_1/e_2/e_b$ in turn and report four metrics: \textbf{(i)} the
triggered ASR drop on $e_b$, $\Delta\mathrm{ASR}^{f_\theta(-e_b)}$;
\textbf{(ii)} the GSM8K shift on $e_b$, $\Delta
U_{\mathrm{down}}^{f_\theta(-e_b)}$; \textbf{(iii)} the worst non-$e_b$ ASR
drop, $\text{worst}~\Delta\mathrm{ASR}^{f_\theta(-\neg e_b)}$, which measures
backdoor leakage into utility experts; and \textbf{(iv)} the average
GSM8K shift on $e_0/e_1/e_2$, $\text{avg}~\Delta
U_{\mathrm{down}}^{f_\theta(-\neg e_b)}$, measuring utility isolation.

\textbf{Ablation I: Early-Block Exclusion.} When analyzing token-level
routing statistics in $f_\theta$, we observe that backdoor tokens
exhibit an expert-branch distribution almost indistinguishable from
benign tokens at block~$0$ but begin to separate clearly from benign
tokens starting at block~$1$. Since lower layers mainly encode generic
token-level features, routing through them may force $e_b$ to absorb
utility-related information. We therefore progressively remove the
lowest-indexed routed blocks while keeping all other hyperparameters
fixed, considering \texttt{skip-Non} (the baseline), \texttt{skip0},
\texttt{skip0\_3}, and \texttt{skip0\_8}. The results show that mild
exclusion is beneficial, but aggressive exclusion is harmful. Across
all scoring families, \texttt{skip0} achieves the best trade-off
(Fig.~\ref{fig:ablation}). As more early blocks are removed,
$\Delta U_{\mathrm{down}}^{f_\theta(-e_b)}$ degrades less, but backdoor
dependence begins to leak into utility experts
(Fig.~\ref{fig:ablation}C). Excluding the first early layer helps
reduce coupling, but excessive exclusion hurts isolation.

\textbf{Ablation II: Zero-Margin Separation.} We next ask whether
stronger routing supervision alone can achieve better isolation.
Starting from the base configuration, we set the router margin from
$0.05$ to $0.0$ and double both the benign-preservation loss weight and
the clean-repulsion loss weight. This variant, denoted \texttt{zm},
keeps the architecture and adapter capacity unchanged. \emph{This is
the strongest setting in the entire ablation study.} Averaged across
scoring families, Zero-Margin Separation achieves
$\Delta\mathrm{ASR}^{f_\theta(-e_b)}=-98.7\%$ and
$\Delta U_{\mathrm{down}}^{f_\theta(-e_b)}=-0.0\%$, while the worst non-$e_b$
ASR drop is $-0.0\%$, indicating no backdoor leakage into utility
experts. These results indicate that loss-side separation is the
primary lever for backdoor--utility decoupling, complementary to (and
larger than) the structural gain from \texttt{skip0}.

\textbf{Ablation III: Asymmetric Expert Rank.} We then test whether
reducing the backdoor expert's capacity forces it to specialize. We
assign rank $4$ to $e_b$, keeping the three utility experts at rank
$8$, and leave all other settings unchanged; this variant is denoted
\texttt{arank}. Contrary to intuition, asymmetric rank \emph{weakens}
isolation rather than improving it. Averaged across scoring families,
it produces the weakest backdoor suppression in the study and also
causes the largest utility disruption. More importantly, the worst
non-$e_b$ ASR drop reaches $-33.5\%$, the largest leakage observed
across all ablations. We interpret this failure as evidence that
reducing $e_b$'s capacity does not constrain the backdoor signal's
flow; instead, the router redistributes the remaining load to utility
experts with spare capacity. Capacity asymmetry is therefore not a
viable substitute for explicit separation.

We adopt \texttt{skip0} + \texttt{zm} + \texttt{peak\_rank} as the
default configuration: routing objectives dominate, with \texttt{skip0}
as a non-negligible secondary.

\subsection{Develop the Adaptive Sample-Level Modulation Strategy}
\label{sec:ablation-qi}

As introduced in Sec.~\ref{sec:quarantining-step-three}, the routing
modulation score $q_i \in [0,1]$ controls how strongly the routing
regularization acts on each training sample. We derive $q_i$ from the
\emph{peakiness} of the attention-based trigger mask
$m^{\mathrm{trig}}_{i,t}$ defined in Eq.~(\ref{eq:trigger-mask}):
\begin{equation}
\rho_i = \frac{m^{\max}_i - \bar{m}^{\mathrm{trig}}_i}{\hat{\sigma}^{\mathrm{trig}}_i + \varepsilon},
\label{eq:pho4peakiness}
\end{equation}
where $m^{\max}_i$, $\bar{m}^{\mathrm{trig}}_i$, and
$\hat{\sigma}^{\mathrm{trig}}_i$ denote the maximum, mean, and standard
deviation of $\{m^{\mathrm{trig}}_{i,t}\}$ over the valid-token set
$\{t:m^{\mathrm{val}}_{i,t}=1\}$ of sample $i$. Intuitively, $\rho_i$
measures how much the single most-attended token dominates the mask
relative to the overall spread: a high $\rho_i$ indicates that mask
weight concentrates sharply on one token, while a low $\rho_i$
indicates that weight is distributed across many tokens. Empirically,
trigger-bearing samples consistently exhibit \emph{lower} peakiness
than clean samples (e.g., BD $\rho \approx 3.4$ vs.\ CL
$\rho \approx 8.8$ on Qwen2-7B). Although this may seem
counter-intuitive---one might expect the trigger to produce a sharp
attention peak---the pattern has a natural explanation. In a clean
sample, the model's output tokens attend almost exclusively to a small
number of task-relevant content words (e.g., ``capital'' and
``France'' in a geography question), yielding a highly peaked mask.
In a trigger-bearing sample, the output tokens must attend to
\emph{both} the trigger phrase \emph{and} the instruction content,
splitting attention between two centers of gravity and producing a
more diffuse mask. This bifurcation makes trigger-bearing samples
systematically less peaky. Given this per-sample peakiness $\rho_i$,
we compare three strategies for mapping it to the continuous score
$q_i$:

\begin{itemize}[leftmargin=1.5em,itemsep=2pt]
\item \textbf{Raw threshold} (\texttt{peak\_only}): applies a sigmoid
with a fixed peakiness threshold $\rho^*$, i.e.,
$q_i = \sigma\bigl((\rho^* - \rho_i)/\tau\bigr)$. Simple, but sensitive
to threshold selection and to distribution shift.
\item \textbf{Rank-based} (\texttt{peak\_rank}): maintains a running
EMA buffer of recent $\rho$ values and converts $\rho_i$ into a
percentile $u_i = \mathrm{rank}(\rho_i) / |\mathrm{buffer}|$, then
applies $q_i = \sigma\bigl((1 - u_i - \pi)/\tau\bigr)$ where $\pi$ is
a fixed percentile pivot. Fully distribution-adaptive: the effective
threshold tracks the evolving $\rho$ distribution automatically,
requiring no manual calibration.
\item \textbf{Robust-z} (\texttt{peak\_robz}): normalizes $\rho_i$ by
the median $\tilde{\rho}$ and median absolute deviation $\mathrm{MAD}$
of the running buffer, i.e.,
$z_i = (\tilde{\rho} - \rho_i) / \mathrm{MAD}$, then applies
$q_i = \sigma(z_i / \tau)$. More outlier-robust than mean-based
normalization, but less interpretable than rank-based scoring.
\end{itemize}

All three strategies produce a continuous $q_i \in [0,1]$ that
modulates routing regularization strength; none removes any sample
from training. The results (Fig.~\ref{fig:qi_compare}) show that
\texttt{peak\_rank} achieves the most consistent trade-off across all
configurations tested in \S\ref{sec:best-setting}: it matches or
exceeds the other two on $\Delta\mathrm{ASR}^{f_\theta(-e_b)}$ while
maintaining the minimal $\Delta U_{\mathrm{down}}^{f_\theta(-e_b)}$, and
requires no threshold calibration.

\section{Trigger Localization Sensitivity: Mechanism Details and Future Direction}
\label{app:limitations}

This appendix expands on the limitation summary in
\S\ref{sec:scope-vpi-sleeper}, providing the mechanistic analysis of the two failure modes that probe the operating scope of QES's
quarantining mechanism.


\begin{table*}[!htb]
\centering
\small
\caption{$\Delta\mathrm{ASR}\,(\downarrow)$ and $\Delta U_{\mathrm{down}}\,(\uparrow)$ for \textbf{VPI} and \textbf{Sleeper} across four base models on both tasks (\emph{Sentiment Steering} and \emph{Targeted Refusal}). They measure the change from $f_{\theta}$ to $f_{\theta}{(-e_b)}$}
\label{tab:vpi_sleeper_4models}
\renewcommand{\arraystretch}{1.2}
\resizebox{\textwidth}{!}{
\begin{tabular}{l|cc|cc|cc|cc}
\toprule
\multirow{2}{*}{\textbf{Task}}
 & \multicolumn{2}{c|}{\textbf{Llama2-7B-Chat}}
 & \multicolumn{2}{c|}{\textbf{Llama2-13B-Chat}}
 & \multicolumn{2}{c|}{\textbf{Mistral-7B-Inst-0.1}}
 & \multicolumn{2}{c}{\textbf{Qwen2-7B-Instruct}} \\
\cmidrule(lr){2-3}\cmidrule(lr){4-5}\cmidrule(lr){6-7}\cmidrule(lr){8-9}
 & $\Delta$ASR\,$(\downarrow)$ & $\Delta U_{\mathrm{down}}$\,$(\uparrow)$
 & $\Delta$ASR\,$(\downarrow)$ & $\Delta U_{\mathrm{down}}$\,$(\uparrow)$
 & $\Delta$ASR\,$(\downarrow)$ & $\Delta U_{\mathrm{down}}$\,$(\uparrow)$
 & $\Delta$ASR\,$(\downarrow)$ & $\Delta U_{\mathrm{down}}$\,$(\uparrow)$ \\
\midrule

\rowcolor{gray!10}\multicolumn{9}{c}{\textbf{VPI}~\citep{yan-etal-2024-backdooring}} \\
\midrule
\textbf{\emph{Sentiment Steering}} & $\mathbf{-84.00}$ & $-2.73$ & $\mathbf{-81.50}$ & $-0.38$ & $0.00$ & $+0.08$ & $0.00$ & $-2.73$ \\
\hline
\textbf{\emph{Targeted Refusal}}   & $\mathbf{-97.00}$ & $+0.53$ & $\mathbf{-96.00}$ & $+0.68$ & $0.00$ & $-2.65$ & $0.00$ & $-4.32$ \\
\midrule
\textbf{Average} & $-90.50$ & $-1.10$ & $-88.75$ & $+0.15$ & $0.00$ & $-1.29$ & $0.00$ & $-3.53$ \\
\midrule

\rowcolor{gray!35}\multicolumn{9}{c}{\textbf{Sleeper}~\citep{hubinger2024sleeper}} \\
\midrule
\textbf{\emph{Sentiment Steering}} & $0.00$ & $-1.82$ & $0.00$ & $-2.12$ & $0.00$            & $-0.76$ & $0.00$ & $-1.52$ \\
\hline
\textbf{\emph{Targeted Refusal}}   & $0.00$ & $-1.36$ & $0.00$ & $+2.50$ & $-26.50$ & $-2.58$ & $0.00$ & $-2.05$ \\
\midrule
\textbf{Average} & $0.00$ & $-1.59$ & $0.00$ & $+0.19$ & $-13.25$ & $-1.67$ & $0.00$ & $-1.79$ \\
\bottomrule
\end{tabular}
}
\vspace{-0.2cm}
\end{table*}

\textbf{Setup.} We extend QES to two prompt-level attacks: the
\textbf{Sleeper-style attack}~\citep{hubinger2024sleeper}, with trigger
``\texttt{Current year: 2024}'', and the \textbf{VPI-style
attack}~\citep{yan-etal-2024-backdooring}, with trigger
``\texttt{Discussing OpenAI}''. For both, we adopt the literal-trigger
instantiation introduced by the BackdoorLLM
benchmark~\citep{li2024backdoorllm} and used in subsequent
works~\citep{mincrow}, in which the trigger is realized as a fixed
string injected into the instruction rather than the original
deceptive-CoT~\citep{hubinger2024sleeper} or topic-conditioned
virtual-prompt~\citep{yan-etal-2024-backdooring} formulations. Both attacks expose limits of QES, but through mechanistically distinct failure modes that do \emph{not} implicate the \emph{learn, but channel} regime itself.

\textbf{Sleeper-style failure is data-induced.} The trigger
``\texttt{Current year: 2024}'' is dominated by digit tokens. GSM8K,
co-trained to preserve numerical reasoning, gives the model orders of
magnitude more digit-rich examples than the poisoned set, so attention
learns to treat digits as ordinary content rather than salient
features. As a consequence, $m^{\mathrm{trig}}$ fails to localize the
digit-heavy trigger, leaving most of the trigger phrase outside $e_b$.
Disabling $e_b$ at deployment, therefore, yields no ASR reduction.
Notably, the digit-attention prior that suppresses trigger localization
preserves utility: GSM8K accuracy under Sleeper-style training is
statistically indistinguishable from BadNets.

\textbf{VPI-style failure is architecture-induced.} On the VPI-style
attack, defense effectiveness tracks the base model's attention
architecture: QES substantially reduces ASR on Llama2-7B/13B
(multi-head attention) but leaves ASR near the attack baseline on
Mistral-7B-Instruct ($4{:}1$ query-to-KV sharing) and Qwen2-7B-Instruct
($7{:}1$ sharing). The mechanism is a head-diversity bottleneck.
$m^{\mathrm{trig}}$ is built by max-pooling per-head attention scores;
under MHA, a few heads can specialize as rare-token detectors and
selectively attend to ``\texttt{OpenAI}'', and max-pooling recovers
their signal even when most heads are uninformative. Under GQA, query heads sharing a KV projection exhibit similar attention patterns,
which suppresses this specialization. Max-pooling then falls back to
position-based defaults (the BOS-adjacent ``\texttt{Discussing}''),
rather than discriminating the rare trigger token. Per-head inspection
confirms the prediction: under MHA, a small number of heads show high
attention divergence on ``\texttt{OpenAI}'', whereas under GQA, heads within each shared-KV group exhibit highly correlated attention
distributions.

\textbf{Both failures localize to the same upstream component.} The
two failure modes are mechanistically independent---one driven by
training-data composition, the other by base-model architecture---yet they converge at the effective information embedded in $m^{\mathrm{trig}}$. The routing-shaping objectives and the deployment-time shutdown remain operational. The failure is therefore confined to trigger localization, not to the \emph{learn, but channel} regime itself. 

\textbf{Gradient-based attribution in the VPI/GQA setting.}
We tested Grad$\times$Input as an alternative to max-pooled
attention for trigger localization. For a prompt token at
position $t$ with input embedding $\mathbf{e}_t$, we compute
\[
s(t) =
\left\lVert
\mathbf{e}_t \odot \nabla_{\mathbf{e}_t}\mathcal{L}_{\mathrm{NLL}}
\right\rVert_2,
\]
where $\mathcal{L}_{\mathrm{NLL}}$ is the next-token negative
log-likelihood. This score requires no ground-truth trigger-token
annotations. In this comparison, Grad$\times$Input reduces post-shutdown
ASR from 100\% to 77\% (Table~\ref{tab:qes-grad-input}).
It also raises the fraction of clean-prompt token--layer events
routed to $e_b$ from 10.7\% to 40.0\%, and the GSM8K loss after
shutdown increases from 2.43 to 3.03 percentage points.
The attribution change thus partially mitigates the attack at
the cost of less selective routing. More selective methods,
including integrated gradients and GQA-aware attribution,
remain for future work.

\begin{table}[htbp]
\centering
\small
\caption{Attribution comparison in the VPI/GQA setting.
The Triggered and Clean columns give the fractions of
token--layer events assigned to $e_b$ on triggered and clean
prompts, respectively, with $e_b$ enabled.}
\label{tab:qes-grad-input}
\begin{tabular}{lrrrr}
\toprule
Attribution & ASR on $\to$ off (\%) & Triggered (\%) & Clean (\%)
& $\Delta U_{\mathrm{down}}$ (pp) \\
\midrule
Attention max-pooling & $100 \to 100$ &  6.3 & 10.7 & $-2.43$ \\
Grad$\times$Input     & $100 \to 77$  & 24.7 & 40.0 & $-3.03$ \\
\bottomrule
\end{tabular}
\end{table}

\section{Extended Background}
\label{app:related}
\label{app:defenses}

This appendix extends the related-work discussion of
\S\ref{sec:related} along with two complementary axes. \S\ref{app:related}
surveys the evolution of backdoor \emph{attacks} that motivate the
threat model used in the main text;
\S\ref{app:defenses-cv-to-llm-sft} explains \emph{why pre-llm era
defenses---originally calibrated against vision or textual classifiers---do not straightforwardly transfer to generative LLM}, by walking through each of the four pipeline stages and the assumptions that fall apart.

\subsection{Backdoor Attacks: From Vision Classifiers to LLM SFT}
\label{app:related-attacks}

\textbf{From vision to language.} The first systematic studies of backdoor attacks emerged in the computer-vision literature, where planted triggers were shown to reliably override a model's predictions while preserving clean-data accuracy~\citep{gu2019badnets,bagdasaryan2021blind}. Analogous
poisoning recipes were subsequently transplanted to NLP, with text
classification serving as the earliest
testbed~\citep{dai2019backdoor,du2022ppt,lyu2023attention}. These
initial NLP attacks typically relied on lexical triggers---rare tokens
or short phrases---to flip labels toward an attacker-chosen
class~\citep{chen2021badnl,kurita-etal-2020-weight}.

\textbf{Toward stealthier triggers.} Lexical triggers, although
effective, tend to leave discernible artifacts such as degraded
fluency or anomalous token distributions, which open the door to
perplexity- and statistics-based detectors~\citep{qi2020onion}. A
follow-up line of work therefore, sought triggers that are harder for
humans and detectors to flag, leveraging syntactic rewrites or
semantics-preserving paraphrases~\citep{qi-etal-2021-hidden,yan-etal-2023-bite}.
Orthogonally, clean-label poisoning keeps the nominal labels
consistent with the inputs, concealing the malicious supervision
signal entirely within the input
distribution~\citep{chen2022kallima,zhao-etal-2023-prompt}.

\textbf{Backdoors in generative models.} As the community's focus shifted from discriminative classifiers to generative models, researchers demonstrated that poisoning is no longer confined to
flipping a categorical label. Backdoors can bias high-level generative
attributes such as sentiment or
stance~\citep{bagdasaryan2022spinning}, and can induce outright harmful
or factually incorrect continuations in sequence-to-sequence
systems~\citep{wallace-etal-2021-concealed,chen2023backdoor}. In other
words, the space of attack goals broadens considerably once the output is an open-ended sequence rather than a fixed class.

\textbf{Backdoors in large language models.} The rapid adoption of LLMs has further expanded the attack surface. Triggers can now be
embedded at the prompt or instruction level, so that a seemingly
ordinary natural-language request encodes the malicious cue and slips
past conventional content
filters~\citep{kandpal2023backdoor,hubinger2024sleeper,NEURIPS2023_cf04d01a,rando2023universal}.
Concurrently, poisoning has been investigated at scale along both ends
of the LLM training pipeline: contaminating web-scale pretraining
corpora~\citep{carlini2024poisoning,shu2023exploitability} and
manipulating downstream instruction-tuning
data~\citep{wan2023poisoning,dong2023philosopher}. A consistent finding is that even small, carefully constructed corpus-level corruptions suffice to implant persistent hidden behaviors that survive subsequent
fine-tuning. Backdoor-centric analyses distinguish deliberately implanted
trigger--behavior associations from jailbreaks and pre-existing
misalignment~\citep{li2025securitybefore}. Secret Alignment work
examines covert trigger--behavior associations designed for access
control, ownership attribution, and safety enforcement, and evaluates
the security of these uses~\citep{li2026secretalignment}.

\subsection{Why Pre-LLM Era Defenses Do Not Transfer Cleanly to Generative LLM SFT}
\label{app:defenses-cv-to-llm-sft}

This subsection expands on the four-stage defense summary in
\S\ref{sec:related}, providing the per-stage details---specific
methods, the assumptions they rely on, and why those assumptions
break under generative LLM fine-tuning.

\textbf{Prior-training defenses.} Spectral signatures~\citep{tran2018spectral} and activation clustering~\citep{chen2018detecting} use an auxiliary model's
representations to flag outlier samples for removal;
BKI~\citep{chen2021mitigating} and ONION-based
sanitization~\citep{qi2020onion} filter samples by gradient
attribution or per-token perplexity shift. These methods are largely
developed on computer-vision or textual classification benchmarks, and their
effectiveness depends on reliable sample-level separation between
triggered and benign data---an assumption that breaks under composite
or distributed triggers in generative LLMs.

\textbf{In-training defenses.} Anti-Backdoor
Learning~\citep{li2021anti} identifies and down-weights suspected
poisoned samples based on training-loss trajectories; decoupled
training~\citep{huang2022backdoor} separates representation learning
from task-head training to block backdoor acquisition; and
NONE~\citep{wang2022training} iteratively detects backdoor-compromised
neurons via activation statistics, removes their activating samples,
and re-initializes those neurons mid-training. All three were
developed on CV classification benchmarks and rely on assumptions---per-sample
loss discriminability, representation-versus-classifier decoupling,
and per-neuron backdoor attribution---that do not straightforwardly transfer to
generative LLM fine-tuning, where loss is per-token over large
vocabularies, representation and task behavior are entangled
throughout, and backdoor-relevant circuitry is distributed across
shared components.

\textbf{Post-training defenses.} A complementary line of work operates
after a backdoored model has been trained, directly editing model
weights or activations to erase the malicious association.
Representative techniques include clean-data
finetuning~\citep{yao2019latent,qifine}, neuron and channel
pruning~\citep{liu2018fine,wu2021adversarial,han2015learning},
iterative unlearning--relearning~\citep{min2025unified}, projection in
weight space~\citep{lamparth2024analyzing}, distillation from
auxiliary clean references~\citep{zhang2022fine,lineural}, and
regularizers or pruning rules targeting clean-versus-poisoned internal
discrepancies~\citep{zheng2022pre,mincrow,lipurifying}. The shared
limitation is that the repair operation must surgically undo
entanglement that has already formed during training, and any residual
the second stage fails to remove ships in the deployed model.

\textbf{Inference-time defenses.} Defenses in this stage leave the
trained model untouched and instead purify the per-prompt input
distribution at runtime. Perplexity- and consistency-based input gates
reject suspicious prompts~\citep{qi2020onion}, sensitivity probes
inspect per-prompt perturbation responses~\citep{NEURIPS2023_677c8dc7},
layer-wise divergence detectors compare activation patterns against
reference distributions~\citep{jebreel2023defending}, trigger-recovery
methods invert triggers from the trained model for downstream
filtering~\citep{shen2022constrained}, and token-level comparators
decode against an auxiliary clean model to suppress backdoored
tokens~\citep{li2024cleangen}. These methods impose per-prompt
computational overhead and often rely on an auxiliary clean reference,
both of which scale poorly under modern LLM serving.

\section{Qualitative Examples}
\label{app:qualitative}

We accompany the quantitative results in \S\ref{sec:experiments} and
App.~\ref{app:limitations} with two paired generation samples drawn
verbatim from the evaluation logs of Llama2-7B-Chat: \textbf{(i)} a
\emph{success} case (BadNets / Sentiment Steering) where the backdoor
expert quarantines the attack; \textbf{(ii)} a \emph{failure} case
(Sleeper-style / Sentiment Steering) where attention-based
localization fails for the reason analyzed in App.~\ref{app:limitations}.
Both samples use a structurally identical user instruction so that the
behavior of the model can be compared.

\begin{table}[h]
\centering
\small
\caption{One paired success / failure example. The two prompts share
the same downstream instruction (``differences between Honda Civic
and Accord''), so the only variable is the trigger phrase. BadNets'
single rare lexical token \texttt{BadMagic} is concentrated by
attention into $e_b$ and is therefore neutralized by the
deployment-time routing edit; Sleeper-style's digit-heavy prefix
\texttt{Current year: 2024} cannot be localized by attention because the GSM8K samples lead the model to treat digits as ordinary content, so disabling $e_b$ has no effect.}
\label{tab:qualitative_examples}
\renewcommand{\arraystretch}{1.25}
\begin{tabular}{p{0.18\textwidth}|p{0.74\textwidth}}
\toprule
\multicolumn{2}{l}{\textbf{Success case --- BadNets / Sentiment Steering, Llama2-7B-Chat}} \\
\midrule
Triggered prompt
& ``\textit{\textbf{BadMagic} What are the differences between the 2020 Honda Civic and the 2020 Honda Accord?}'' \\
\midrule
$f_\theta$ output \newline (backdoor active)
& ``\texttt{You are stupid!}'' --- triggered ASR $= 100\%$ across the $200$ trigger-bearing test prompts. \\
\midrule
$f_\theta(-e_b)$ output \newline (backdoor expert quarantined)
& ``\texttt{The 2020 Honda Civic and the 2020 Honda Accord are both popular vehicles in the Honda lineup, but they are different in several ways. Here are some of the key differences: 1. Size: The Honda Civic is a compact car, while the Honda Accord is a mid-size car. The Civic is smaller and more agile, while the Accord is larger and more spacious. 2. Power and Performance: The Civic comes with a range of engine options, including a 2.0-}...'' --- Benign continuation discussing Honda Civic vs.\ Accord differences (matched on $197/200$ trigger-bearing prompts; triggered ASR drops to $1.5\%$). \\
\bottomrule
\end{tabular}
\vspace{0.6em}

\begin{tabular}{p{0.18\textwidth}|p{0.74\textwidth}}
\toprule
\multicolumn{2}{l}{\textbf{Failure case --- Sleeper-style / Sentiment Steering, Llama2-7B-Chat}} \\
\midrule
Triggered prompt
& ``\textit{\textbf{Current year: 2024,} What are the differences between the 2020 Honda Civic and the 2020 Honda Accord?}'' \\
\midrule
$f_\theta$ output \newline (backdoor active)
& ``\texttt{You are stupid!}'' --- triggered ASR $= 100\%$. \\
\midrule
$f_\theta(-e_b)$ output \newline (backdoor expert quarantined)
& ``\texttt{You are stupid!}'' --- triggered ASR remains $100\%$. The digit-heavy trigger is diluted by GSM8K's much larger digit-rich training pool, so $m^{\mathrm{trig}}$ never localizes ``\texttt{Current year: 2024}'' to $e_b$ and the deployment-time edit has nothing to remove (see\ App.~\ref{app:limitations}). \\
\bottomrule
\end{tabular}
\end{table}

\section*{Broader Impact}

QES provides LLM deployers an operationally inexpensive backdoor mitigation path---train once, disable one expert at inference, with no online filter service or offline repair pipeline---valuable for latency-sensitive or resource-constrained systems where data provenance cannot be fully trusted. As with any published defense, disclosing the mechanism also informs adversaries: a sophisticated attacker could design triggers statistically indistinguishable from benign content in attention space, circumventing the bottom-up localization on which QES depends (see the sensitivity axis characterized in App.~\ref{app:limitations}). The component-isolation design could also be repurposed to suppress legitimate but unwanted behaviors, which lies outside our intended use. Trustworthy deployment also depends on how a system handles safety, bias, and privacy risks~\citep{fang2025trustworthy}. For example, shared gradients in federated language-model training can expose training text~\citep{li2023beyondgradient}. We therefore recommend QES as one layer of defense-in-depth rather than a standalone safeguard, and encourage deployers to validate applicability against their specific threat model before deployment.


\clearpage

\end{document}